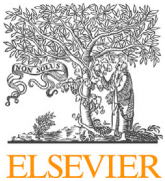
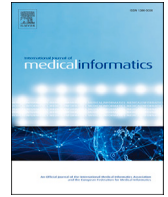

Review article

# Sensing technologies and machine learning methods for emotion recognition in autism: Systematic review

Oresti Banos [a,*], Zhoe Comas-González [a,b], Javier Medina [a], Aurora Polo-Rodríguez [a,c], David Gil [d], Jesús Peral [e,*], Sandra Amador [d], Claudia Villalonga [a]

[a] *Department of Computer Engineering, Automation and Robotics, University of Granada, Granada, Spain*
[b] *Department of Computer Science and Electronics, Universidad de la Costa, Barranquilla, Colombia*
[c] *Department of Computer Science, University of Jaén, Jaén, Spain*
[d] *Department of Computer Technology and Computation, University of Alicante, Alicante, Spain*
[e] *Department of Sotware and Computing Systems, University of Alicante, Alicante, Spain*

ARTICLE INFO

ABSTRACT

*Background:* Human Emotion Recognition (HER) has been a popular field of study in the past years. Despite the great progresses made so far, relatively little attention has been paid to the use of HER in autism. People with autism are known to face problems with daily social communication and the prototypical interpretation of emotional responses, which are most frequently exerted via facial expressions. This poses significant practical challenges to the application of regular HER systems, which are normally developed for and by neurotypical people.
*Objective:* This study reviews the literature on the use of HER systems in autism, particularly with respect to sensing technologies and machine learning methods, as to identify existing barriers and possible future directions.
*Methods:* We conducted a systematic review of articles published between January 2011 and June 2023 according to the 2020 PRISMA guidelines. Manuscripts were identified through searching Web of Science and Scopus databases. Manuscripts were included when related to emotion recognition, used sensors and machine learning techniques, and involved children with autism, young, or adults.
*Results:* The search yielded 346 articles. A total of 65 publications met the eligibility criteria and were included in the review.
*Conclusions:* Studies predominantly used facial expression techniques as the emotion recognition method. Consequently, video cameras were the most widely used devices across studies, although a growing trend in the use of physiological sensors was observed lately. Happiness, sadness, anger, fear, disgust, and surprise were most frequently addressed. Classical supervised machine learning techniques were primarily used at the expense of unsupervised approaches or more recent deep learning models. Studies focused on autism in a broad sense but limited efforts have been directed towards more specific disorders of the spectrum. Privacy or security issues were seldom addressed, and if so, at a rather insufficient level of detail.

## 1. Introduction

Autism spectrum disorder (ASD) is a neurodevelopmental condition characterized by a deficit in communication, social interaction, and lack of understanding of emotions. It affects circa 1% of the population and can be detected in the first years of life [1]. One of the key reasons for the emotional misunderstanding is the inability of people with autism to comprehend prototypical feelings and emotions, which directly affects social interaction. In view of this challenge, some research has been lately devoted to the automatic recognition of human emotions in autism. This research area is largely based on the well-established field of Human Emotion Recognition (HER), which exploits sophisticated sensing technologies and advanced machine learning techniques to detect and understand the feelings and emotions of people. Even when HER systems have been used to detect, intervene, and accompany the adaptation process of people with autism into society, it is generally accepted that existing approaches are not definitive. Many of these studies deal with biased data and recognition of emotions such as happiness or fear was only marginally impaired in autism as well as the generalizability of the findings from the currently available data remains unclear [2,3]. Furthermore, HER algorithms primarily rely on facial cues, overlooking other important aspects such as body language, vocal tone, and contextual and situational factors that would improve the accuracy of the algorithms [4]. [5] and [6] describe new tools as well as computational model to assist people with autism in understanding and operating in the socioemotional world around them. Some findings reveal that children with autism spectrum condition have residual difficulties in this aspect of empathy. In the work of [7] authors concluded that relations between particular emotions and human body reactions have long been known, but there remain many uncertainties in selecting measurement and data analysis methods Moreover, it is also observed that a great number of the HER models used in autism are based on data collected from neurotypical people [8]. Be that as it may, the use of general HER models in autism-related applications poses a number of challenges yet to be addressed and which demand special attention from the scientific community. While there exists a great bulk of systematic reviews addressing the technologies and methods used for emotion recognition in general [7,9–11], very few focus specifically on its use in autism. In fact, existing systematic reviews in this direction are either centred on a specific technology such as eye-tracking [12], robots [13], and wearables [14], or particular methods like deep learning [15]. Hence, a comprehensive systematic review focusing on the state of the art on emotion recognition sensing technologies and machine learning methods for autism emotion recognition is presented here. The results of this review will contribute to improve the current techniques for emotion recognition used in autism studies, encourage new research focusing on other conditions of the autism spectrum disorder that have been marginally investigated to date, and promote the use of physiological methods in addition to other traditional behavioural methods as potential emotion recognition modalities to be used in autism. The primary objective of this review was to determine the trends, advances, and challenges on sensing technologies and machine learning methods for emotion recognition in autism. To that end, this review aimed to answer the following research questions: (1) What type of sensor technology has been used for emotion recognition in autism?; (2) What type of machine learning techniques are most commonly used for emotion recognition in people with autism?; and (3) What are the main challenges in the use of emotion recognition technologies in people with autism? To the best of our knowledge, there are many reviews on autism, on HER, on machine learning methods but very little written about the whole of them and their complementation of these different areas. This is the main novelty of this review. Our study covers all age groups unlike most studies that focus on children. We raised a specific question to identify the main challenges in the use of emotion recognition technologies in autism. We also provide privacy and security aspects including the use of inform consents or approval by ethics committees. Furthermore, we offer a more recent view on the art as its search reaches up to June 2023.

## 2. Methods

The PRISMA 2020 (Preferred Reporting Items for Systematic Reviews and Meta-Analyses) guidelines [16] were followed to perform a systematic review of the literature on sensing technologies and machine learning methods for emotion recognition in autism. The specific methodology followed is described in the following sections.

### 2.1. Eligibility criteria

This review focused on studies that dealt with sensor technology and machine learning techniques for emotion recognition in children, young, and adults with autism. We did not restrict study location, sample size, gender, age, autism type, type of emotion, emotion recognition modality, devices and sensors, nor algorithms. Studies were eligible to be included in this review if they had three characteristics: 1) they were related to emotion recognition; 2) used sensors and machine learning techniques; and 3) involved children with autism, young or adults.

Other eligibility criteria included: 1) published between January 2011 and June 2023; 2) written in English; 3) scientific article published in a journal or in conference proceedings; and 4) research domain related to computer science or engineering.

Studies were ineligible if affective technology was used in therapy and treatment of patients with autism or in an educational environment. Therefore, we excluded studies related to: 1) robotic treatments or therapies; and 2) social interaction and education.

### 2.2. Information sources

We conducted electronic searches for eligible studies within the reference databases of Scopus and Web of Science. The search was conducted from 1st January 2011 to 30th June 2023.

### 2.3. Search strategy

"Autism" and "emotion recognition"/"recognition of emotion" were selected as primal concepts to be searched. In addition to them, synonyms of the "autism" term, namely "autistic", and the "emotion" term, namely "mood" and "affect", were also considered as they are quite often used interchangeably in this research area. Limits were also applied to the search strategy based on the eligibility criteria. We selected papers published between 2011-2023, published in English computer science or engineering journals or proceedings. The resulting queries eventually run on Scopus and Web of Science are shown below.

**Scopus:**

TITLE-ABS-KEY(((autism OR autistic) AND ("emotion recognition" OR "mood recognition" OR "affect recognition" OR "recognition of mood" OR "affect recognition" OR "recognition of affect")) AND (LIMIT-TO (PUBYEAR,2023) OR LIMIT-TO (PUBYEAR,2022) OR LIMIT-TO (PUBYEAR,2021) OR LIMIT-TO (PUBYEAR,2020) OR LIMIT-TO (PUBYEAR,2019) OR LIMIT-TO (PUBYEAR,2018) OR LIMIT-TO (PUBYEAR,2017) OR LIMIT-TO (PUBYEAR,2016) OR LIMIT-TO (PUBYEAR,2015) OR LIMIT-TO (PUBYEAR,2014) OR LIMIT-TO (PUBYEAR,2013) OR LIMIT-TO (PUBYEAR,2012) OR LIMIT-TO (PUBYEAR,2011)) AND (LIMIT-TO (LANGUAGE,"English")) AND (LIMIT-TO (DOCTYPE,"ar") OR LIMIT-TO (DOCTYPE,"cp")) AND (LIMIT-TO (SUBJAREA,"COMP") OR LIMIT-TO (SUBJAREA,"ENGI")) AND (LIMIT-TO (SRCTYPE,"p") OR LIMIT-TO (SRCTYPE,"j")))

**Web of Science:**

(TS = ((((autism OR autistic) AND ("emotion recognition" OR "mood recognition" OR "affect recognition" OR "recognition of mood" OR "affect recognition" OR "recognition of affect"))))) AND (PY = = ("2023" OR "2022" OR "2021" OR "2020" OR "2019" OR "2018" OR "2017" OR "2016" OR "2015" OR "2014" OR "2013" OR "2012" OR "2011") AND DT = = ("ARTICLE") AND SJ = = ("ENGINEERING" OR "COMPUTER SCIENCE") AND LA = = ("ENGLISH"))

### 2.4. Selection process

The records retrieved from the databases and hand search were imported to the Mendeley Web Library, which was used as a primary tool to navigate through both records and reports. Duplicate records were manually identified by cross-checking title and abstract and then removed by three reviewers (ZC, OB, CV). These reviewers also screened each record and each report retrieved, assessed their eligibility, and eventually selected the final set of studies to be included in the review after reaching a majority consensus.

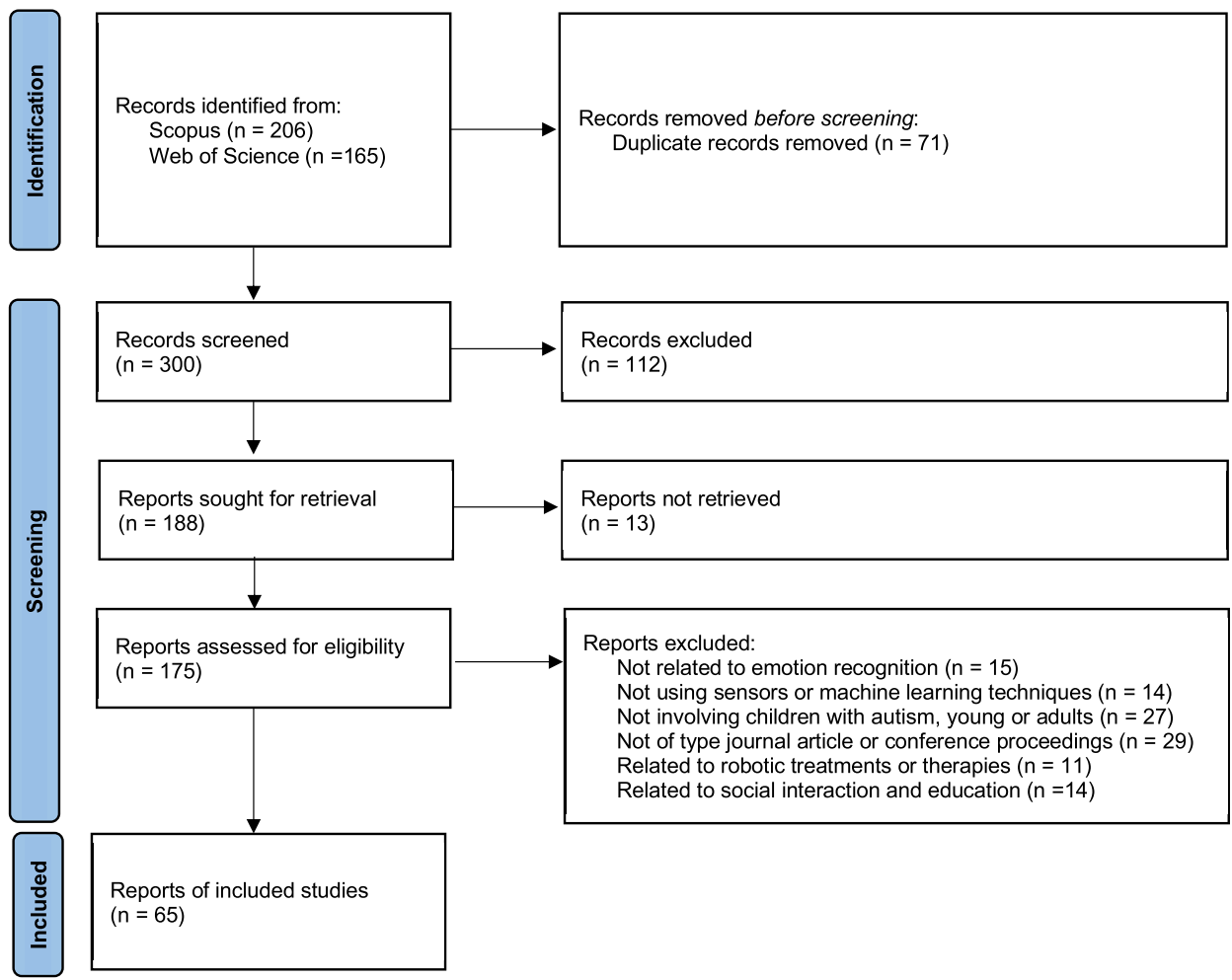

**Fig. 1.** Prisma flowchart.

## 2.5. Data collection process

All reviewers (ZC, OB, JM, AP, DG, JP, SA, CV) participated in the review and assessment of the included studies. The studies were evenly distributed among three groups of reviewers according to their affiliation.

We used a cloud-based collaborative spreadsheet (Google Spreadsheet) to collect data from the included studies. The document consisted of a state-of-the-art matrix where each row represented a study and the columns indicated the data items to be analyzed. Each group of reviewers had to full screen and analyze the papers that were assigned to them and fill the information in the corresponding columns of the matrix. Periodic meetings were held in order to harmonise terminology and overcome potential discrepancies in the assessment process. Reviewers worked independently to extract the information.

## 2.6. Data items

The columns defined in the collaborative spreadsheet corresponded to the outcomes for which data were sought. The specific columns defined were: study name, year of publication, type of article, research goals, subject condition (autism type), emotion recognition modality, dataset (collection or use of), description of the dataset (if applicable), emotions sensed, devices used for the data collection, machine learning techniques, validation methods, study sample (size, type), study length, performance results, study outcomes, privacy and security, and challenges and future work.

## 3. Results

### 3.1. Study selection

A sample of 371 records were identified from the literature search. Namely, the search in Scopus yielded 206 records, while 165 records were obtained for Web of Science. 71 duplicate records were removed before screening. After deduplication, 300 records remained and were screened based on title and abstract. 112 records were excluded and 188 reports were sought for retrieval. 13 reports could not be retrieved and the remaining 175 reports were assessed for eligibility. 110 reports were excluded according to the eligibility criteria and the remaining 65 reports were included for the analysis. The workflow with the detailed process is shown in Fig. 1.

### 3.2. Research goal

The objectives for investigating emotion recognition in autism vary across studies. However, certain common goals are shared among some of these studies. Thus, for example, 29% (19/65) of the studies propose and analyze algorithms and machine learning techniques to automatically recognize emotions in people with autism [17–34]. Around 18% (12/65) of the studies propose the development and application of video games and apps to help children with autism understand and express emotions [8,35–45]. Fewer than 5% (3/65) of the studies explore the use of physiological signals for the automated identification of emotions [19,46,47]. The remainder of the studies have disparate goals. All research goals are detailed in Table A.1.

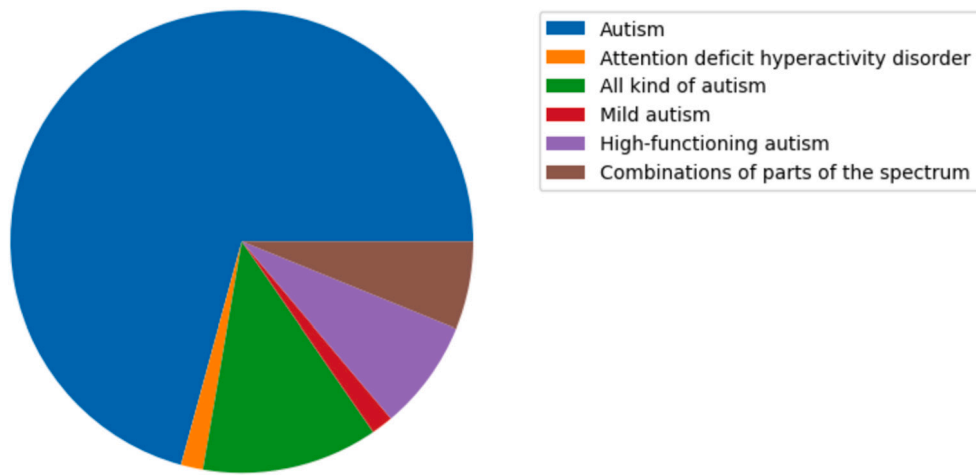

**Fig. 2.** Autism types investigated in each reviewed study. (For interpretation of the colours in the figure(s), the reader is referred to the web version of this article.)

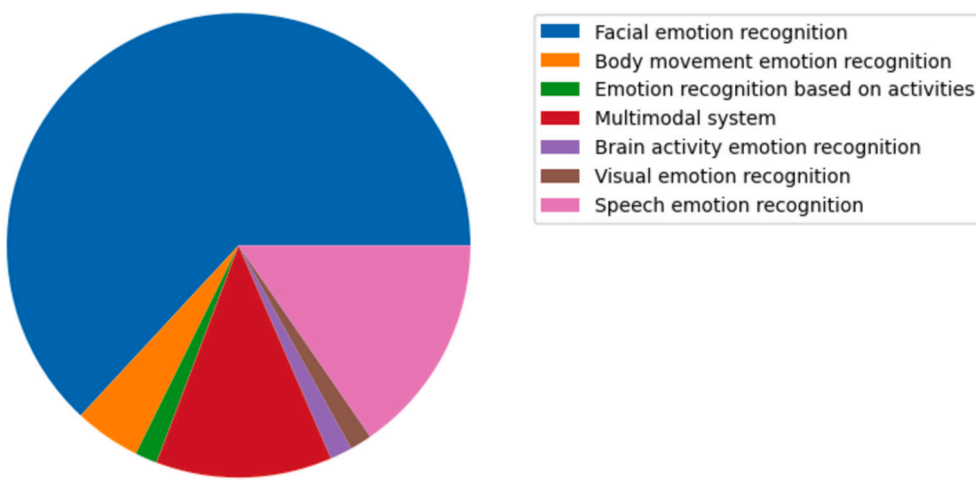

**Fig. 3.** Emotion recognition modality and sensed body regions for each reviewed study.

### 3.3. Autism type

The type of autism varies across studies (Fig. 2). For example, 70% (46/65) of the studies address "autism" in general [8,17–23,28–35,38–40,42,44,46,48–71], while 12% (8/65) refer to the full spectrum as "all kind of autism" [24–27,37,72–74]. Studies referring to "autism" tend to address broad aspects that apply to the overarching spectrum of ASD. In contrast, when some studies specifically mention "all kind of autism", they appear to suggest a deliberate effort to encompass the diverse manifestations and subtypes within the autism spectrum, recognizing and considering the heterogeneity of the condition. Nonetheless, most often both terms are used interchangeably. In some studies specific subtypes of the disorder are addressed. For example, 8% (5/65) correspond to high-functioning autism [45,61,75–77]. Less than 2% (1/65) to mild autism [78] and 2% (1/65) to attention-deficit hyperactivity disorder [79]. 6% (4/65) address combinations of parts of the spectrum, namely middle and moderate autism [47], all kinds of autism and Asperger [36], and high-functioning autism and Asperger [41,80]. While there exist a varied distribution of research focus across subtypes, the contributions in this regard are comparatively low. The limited focus on these subtypes may suggest that the majority of research in ASD aims to address broader aspects of the spectrum rather than delving into detailed examinations of specific subtypes. In Table A.2, the subtype of autism addressed in each selected paper is listed.

### 3.4. Emotional expressions

The selected studies focus on diverse emotional responses, expressions and sensed body regions or signals (Fig. 3). The most common emotion recognition modality is based on facial expressions, accounting for 63% (41/65) of the studies [8,20,22,23,26,28–32,34–38,40–46,49,51–56,58–61,64,67,68,72,75,76,79,80]. 15% (10/65) are based on speech aspects [25,27,63,65,66,69,71,73,74,77], followed by 5% (3/65) exploiting body movement [21,39,48], 2% (1/65) based on daily activities [17], 2% (1/65) measuring brain activity [19], and 2% (1/65) focusing on eye activity [78]. 12% (8/65) of the studies consist of a multimodal approach which combine some of the above [18,24,33,47,50,57,62,70]. This distribution underscores the prevalent reliance on facial expressions while recognizing the significance of speech-related aspects in understanding and studying emotions within ASD.

In relation to the sensed body regions or signals, the majority of studies 71% (46/65) use physical data, i.e. sensed from the external parts of the body, mostly the face. 20% (13/65) of the studies exploit the inner body, including physiological signals such as electroencephalography (EEG) and electromyography (EMG), or psychoacoustic signals [19,24,25,27,47,63,65,66,69,71,73,74,77]. The neurophysiological approaches provide valuable insights into the neural and muscular correlates of emotional states. As for the rationale behind considering psychoacoustics lies in the fundamental role of voice in the recognition of emotions within human interactions. By delving into the nuances of voice expression, researchers aim to deepen their understanding of how emotions are conveyed and perceived through auditory cues, con-tributing to a more comprehensive exploration of emotional recognition within the context of ASD. Around 8% (5/65) combine both physical and physiological signals [33,35,50,62,70]. Less than 2% (1/65) of the studies did not provide enough information to this respect [17] (Table A.3).

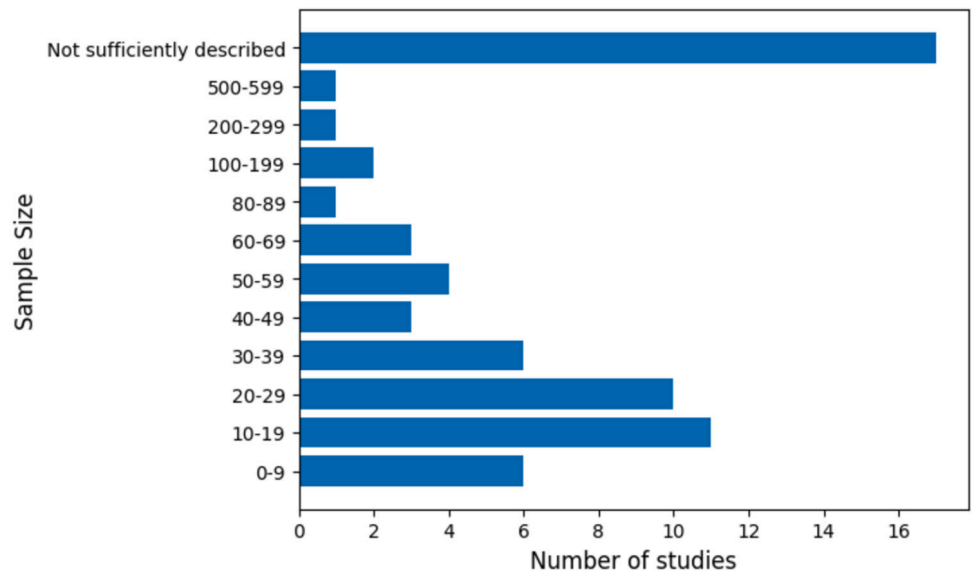

**Fig. 4.** Distribution of the reviewed studies according to the sample size.

*3.5. Study characteristics*

The average number of participants was 49, calculated from the 48 studies indicating the number of participants (74% of the studies) [8,18,19,21–27,29,30,32,33,36–41,44–48,50–55,57,58,60–62,65, 67,70,72–80]. The minimum sample size was four subjects [67] and the maximum 500 [22]. This diversity in sample sizes underscores the variability in research approaches within the field, with some studies opting for smaller, more focused samples, while others involve larger cohorts. Seventeen papers did not specify this number [17,20,28,31,34,35,42,43,49,56,59,63,64,66,68,69,71]. The absence of participant count details in a significant number of papers highlights the need for increased transparency and reporting consistency in research methodologies. The distribution of the studies based on the sample size is shown in Fig. 4.

One day was the minimum study duration [19] and 140 days the maximum duration [48]. Yet, it must be noted that no additional information is provided in the rest of studies to this respect. The absence of duration details in the rest of the studies emphasizes the need for improved reporting standards to ensure a comprehensive understanding of the temporal aspects of HER research in autism.

As for the neurodevelopmental disorder distribution, 51% (33/65) of the studies involved people with autism [8,18,21,22,25,27,29,32,33, 37–41,44–47,51,54,57,58,61,62,65,67,70,72,75–78,80]. Almost 28% (18/65) included both people with and without autism [8,18,25,27,29, 38,41,44,45,47,54,57,58,74–79]. This approach allows researchers to explore and understand the unique features associated with ASD by contrasting them with individuals without the disorder. Around 9% (6/65) of the studies considered only people without autism [30,50,52,53,55, 81]. Although it is not always clearly stated, the reasons for only considering neurotypical individuals are either the desire to establish baseline characteristics or more often the lack of access to people with autism. In 32% (21/65) of the studies, the disorder is not precisely described [17,19,20,23,24,26,28,31,35,36,42,43,48,52,56,60,63,64,66,68,73]. In 23% (15/65) an existing dataset was used to test the proposed solution [19,28,30–32,34,40,42,43,49,63,66,68,69,71].

Less than 28% (18/65) of the studies involved subjects of both genders [18,23,24,27,29,30,36,37,39,50,54,57,65,70,72–74,76]. 5% (3/65) of the studies only involved males [46,77,79] while 6% (4/65) only included females [26,52,53,55]. The remaining 59% (38/65) of the studies did not specify the gender of the participants [8,17,19–22,25,28,35,38,47–49,51,56,58–62,75,78,80]. Inadequate reporting of participant gender in these studies hinders the interpretability of outcomes and, more significantly, impedes a thorough analysis of gender-related effects and potential learning biases.

Regarding the age of the involved subjects, 55% (36/65) of the studies provided this value [8,18,21,23–25,27,29,30,36–41,44,46,47, 51,52,54,57,58,60,62,65,67,70,72–80] resulting in an average age of 18±10 years old. This suggests a relatively diverse age range among the participants, which could have implications for the generalizability of the proposed HER systems across different developmental stages. The other 45% (29/65) of the studies did not mention any age details [17,19,20,22,26,28,31–35,42,43,45,48–50,53,55,56,59,61, 63,64,66,68,69,71]. All this information is summarized in Table A.4.

*3.6. Types of emotions*

The set of emotions analyzed in the selected studies are broadly based on the six universal emotional expressions, i.e. "anger", "sadness", "happiness", "disgust", "surprise", and "fear" [82]. 43% (28/65) of the studies focused on these six basic emotions [18,22,23,26,27, 30,32,35,37,38,47–50,56–61,73–80]. Two studies [20,68] used these very six emotions but replacing "disgust" with the "neutral" emotional state. Another study only uses four basic emotions adding "delight" and "joy" emotions [34]. The remaining 33 studies (51%) [8,17,19,21,24, 25,29,31,33,36,39–46,50–55,62–67,69,71,72] used a number of emotions ranging from two to nine primitives. The emotions considered in addition to the six basic ones were "neutral", "calm", "nervous", "scared", "curious", "excited", "sleepy", "contempt", "joy", "interested", "positive", "positive and talking", "odd positive", "negative", "boredom", and "contentment".

Table A.5, Fig. 5 and Fig. 6 show the emotions used in all the analyzed studies for training/validation and test respectively. According to the listed results, approximately half of the studies leverage the six universal emotions, often relying on or producing publicly available datasets accessible to the scientific community. This choice facilitates meaningful comparisons between these studies, given their shared use of a standardized set of emotions. In contrast, the remaining studies opt for or create "ad-hoc" specific datasets, employing a set of emotions distinct from the universal ones. As a result, conducting comparisons be-

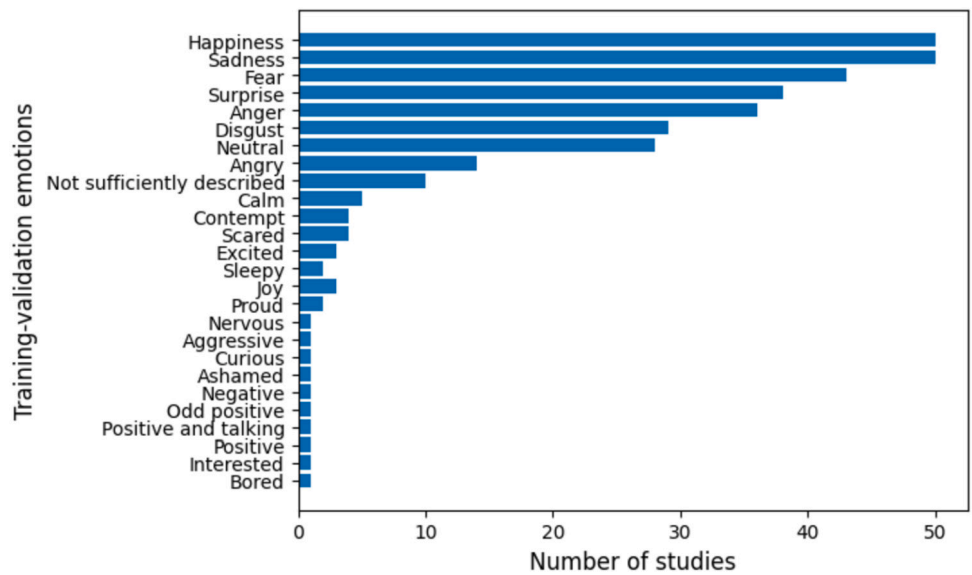

**Fig. 5.** Emotions used for the training-validation of the recognition models in the reviewed studies.

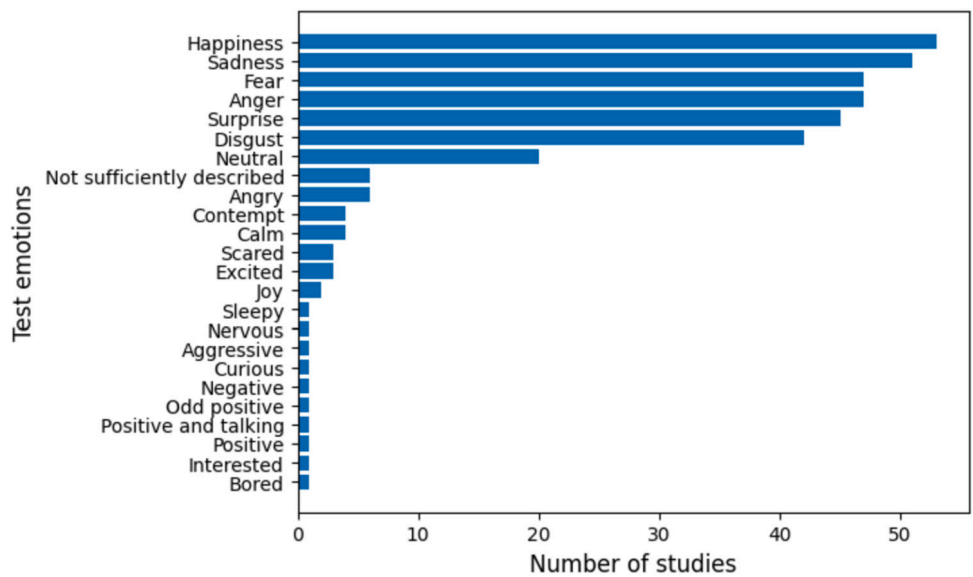

**Fig. 6.** Emotions used for the testing of the recognition models in the reviewed studies.

tween these approaches becomes more intricate due to the varied and specialized nature of the emotional categories used in these datasets.

Although some studies did not mention the emotions used in the training and validation of the HER models [28,29,36,37,44,45,67,70, 75,78], a prevailing trend is the consistent use of the same set of emotions across training, validation, and test phases in most studies. Exceptions to this are [25–27,43,47,58–60,73,74,76,77,80], which used different sets of emotions for training and validation than for test, representing 20% (13/65) of the studies. Employing a different set of emotions for test introduces valuable diversity, reflecting the model's adaptability to recognize a broader spectrum of emotional expressions beyond its training data. This approach enhances the robustness and real-world applicability of HER models by challenging them with unseen emotion data instances during evaluation.

*3.7. Devices and sensors*

Two generations of devices and sensors are identified for the time frame considered for this review, which is related to the periods 2011-2014 and 2015-2023, respectively.

Around 11% (7/65) of the studies correspond to the period 2011-2014, which is characterized by using images and audio as the primary data source. 57% (4/7) of these studies use a so-called first generation of devices consisting of webcams, headphones, and microphones [36,73,74,77]. To facilitate the labelling of the user's data, some controls were incorporated into the systems in 43% (3/7) of the aforementioned studies, including control knobs [74,77] or numeric keypads [79], which were easily handled by users with autism. This period marked the initial steps in using technology for autism research, establishing a foundation for future studies.

Circa 89% (58/65) of the studies represent the period 2015-2023, which is distinguished by a second generation of more advanced and ubiquitous devices and sensors. The technological advancement introduced a wide range of versatile devices, including mobile phones, tablets, 3D cameras, infrared cameras, and more, expanding the capabilities for data collection. Thus, for example, handheld devices are included in 16% (9/58) of these studies, including mobile phones [30,40,50], tablets [37,78], and other handheld devices [58]. Some of these devices were used to support gamification apps [8] or to exploit the mobile camera sensor for recognition purposes [30,35].

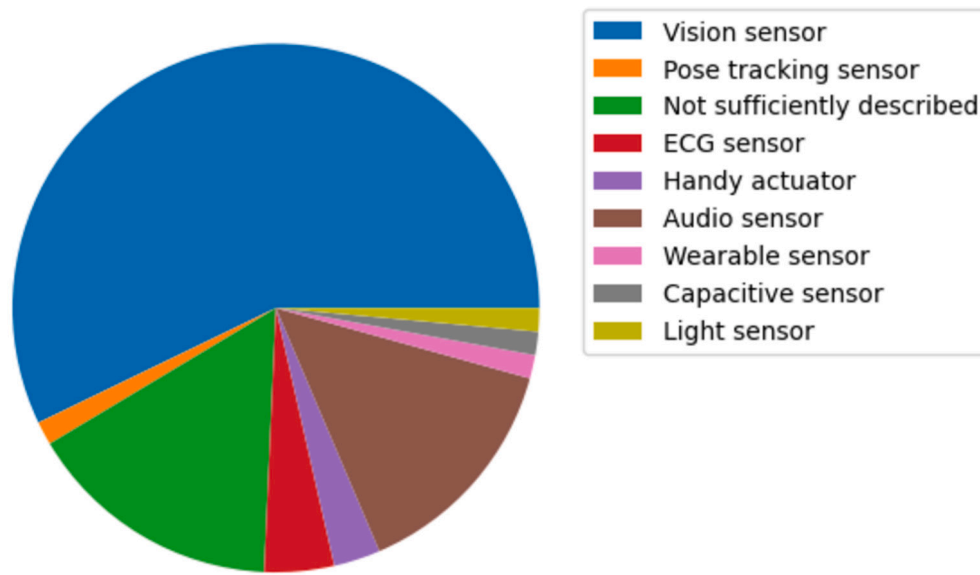

**Fig. 7.** Devices, sensors, and specific models used in the reviewed studies.

In particular, the new generation of cameras was used in 7% (4/58) of the studies, including IP cameras [29,56], 3D cameras [51], and infrared cameras [76]. However, classic vision and audio devices and sensors continued to be used in 33% (19/58) of the studies [20,21,25,38–40,43,54,59,60,62–67,70,71], most likely due to the prevalence of the study of physical cues in autism research.

The advent of facial and body tracking technologies was also leveraged in this field. Such technologies were used in 16% (9/58) of the studies of the so-defined second generation. Devices like Kinect and Intel RealSense enabled improved facial and body tracking, enhancing the interaction and analysis of autistic behaviours. Kinect devices were incorporated into various works [17,45,48,56] due to the availability of an RGB camera, a depth sensor, and a microphone of-the-shelf. Recently, some works have started to use the Intel RealSense device, which has characteristics similar to Kinect [41]. In the same way, Tobii devices were proposed for eye tracking [57] or for head and eye tracking [18]. Standard cameras were also used to record images for eye tracking [75] and pose tracking [48].

Physiological sensors such as EEG were incorporated in 5% (3/58) of the second generation studies. In [19,47] EEG data is collected using a headset with electrodes placed on the participants' scalps. In [46], the authors use a commercial EEG device (Emotiv) to collect data from the frontal, temporal, and posterior brain regions. The use of wearable devices was rare. Only 3% (2/58) of the studies included these devices. Microsoft Band 2 was used in [56], while shimmer sensors were used in [47]. In an attempt to incorporate augmented reality features, Microsoft Hololens and Google Glasses were also used in [52,54], respectively. Full details are provided in Table A.6 and Fig. 7.

### 3.8. Models and performance

This subsection summarises the findings concerning machine learning techniques, performance and metrics, validation methods, and the number of data samples.

The reviewed studies make primary use of supervised learning techniques. Support vector machines (SVM) stand out as the most widely used technique, namely in 29% (19/65) of the studies [8,19,21,23,24, 27,29,31,39,41,46–50,60,69,73,74]. SVM, together with other classical machine learning techniques such as decision trees (DT), random forest (RF), logistic regression (LR) and nearest neighbours k (KNN), are present in roughly 54% (35/65) of the studies reviewed. The remaining 17% (11/65) of the studies do not provide any information on the techniques used [36,37,56–59,61,75,78–80].

The use of Deep Learning is currently confined to recent works, constituting a 31% (20/65) of the studies [20,25,28–34,40–43,63–66,68, 69,71]. This limitation suggests an untapped opportunity, as earlier research may not have fully harnessed the capabilities of deep learning for complex pattern recognition tasks in emotion recognition in autism. Notably, some of the most recent works leverage Deep Learning techniques, including convolutional neural networks, highlighting the emerging potential for improved performance in emotion recognition. However, it is crucial to acknowledge a potential bias towards supervised learning, indicating a potential gap in exploring unsupervised or semi-supervised methods. These alternative approaches could offer valuable insights, especially in scenarios where labelled data is scarce or challenging to acquire. Exploring a broader spectrum of deep learning methodologies could enhance the versatility and effectiveness of emotion recognition models.

The performance of emotion recognition models varies significantly among the studies, attributed to differences in target emotions, sensor data types, machine learning techniques, and dataset instances. This variation suggests challenges in directly comparing study outcomes and establishing standardized benchmarks. Studies can be categorized into three groups according to performance levels. First, 28% (18/65) of the studies are ranked as of high performance (i.e. accuracies greater than 90%), most usually developing an offline evaluation based on datasets collected under controlled conditions [8,21,23,24,26,28,32,40,41,43, 49,55,59,67,70,72,76,80]. The second group characterizes by performances within the range 80-90%, representing 23% (15/65) of the studies [17,18,22,29,38,42,48,53,58,60,63,66,73,74,78]. The third and last group encompass 26% (17/65) of the studies [19,20,25,31,33,39,40, 50–52,54,62,64,65,68,69,77], with more ambitious and challenging solutions based on emerging sensor technologies, leading to performances below 80%. The remaining studies have not sufficiently described their performance results and could not be classified into any group.

The studies make use of a variety of metrics to evaluate model performance, including accuracy, sensitivity, and specificity. This range of metrics provides a fair understanding of model performance, especially those handling dataset imbalances. Concerning the metrics used to estimate model performance, around 57% (37/65) of the studies have chosen the use of accuracy [8,17–24,28–30,32,33,35,38–43,46,48–50, 52–55,62–64,66–70]. Other studies have used unweighted average recall to deal with data set imbalance more effectively [27,30,65,74]. Sensitivity has also been used in some studies [51,59,72], although they still need to evaluate the prominence of true negatives by avoiding the use of specificity. Few studies [18,22,41] rely on the use of accuracy,

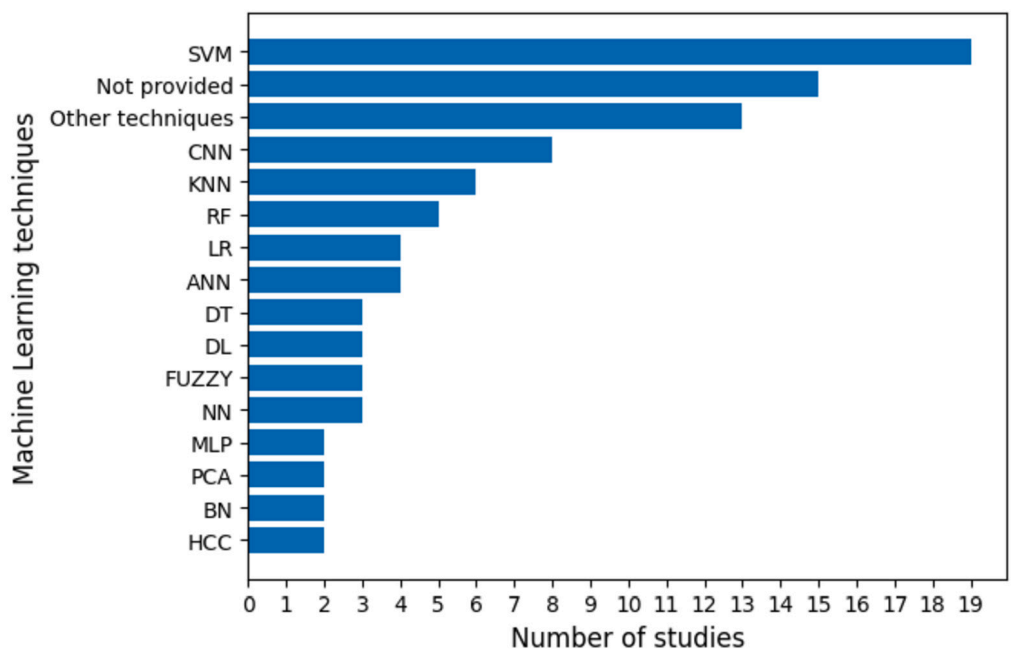

**Fig. 8.** Machine learning techniques, performance and metrics, and validation methods used in each of the reviewed studies.

sensitivity, and specificity to more fully reflect the performance of their recognition system.

The studies adopted different cross-validation techniques (e.g., ten-fold, five-fold, leave-one-out) which provide a rigorous approach to model validation, ensuring the reliability of the findings. Cross-validation is developed in 22% (14/65) of the studies: ten-fold cross-validation is used in [19,21,29,49,53,54,60], five-fold cross-validation [31,33,46], eight-fold cross-validation [46] and one-leave-out cross-validation [18,28,74]. More exceptionally, other approaches such as random split leave one subject out [48], random split cross validation [32,52], and split train-test or hold out [26,38,42,43,63,68,69] are used. A more detailed description can be found in Table A.7 and Fig. 8. This figure illustrates the variety of machine learning methods employed in the reviewed papers, emphasizing the growing prevalence of deep learning due to its robust yet intricate models. However, a noteworthy 12% (8/65) of the papers lack information on the techniques utilized. Encouragingly, there is an expectation that this trend will shift, leading to more papers sharing their models' code in repositories for enhanced scientific community knowledge.

*3.9. Information privacy and security*

Despite the relevance of ensuring privacy and security policies in this field, only 22% (14/65) of the studies acknowledge these sufficiently [18,23,24,29,36,39,41,47,54,58,61,67,72,79]. Three of these studies [23,72,79] followed the 1964 Declaration of Helsinki, a formal statement of ethical principles published by the World Medical Association (WMA) to guide the protection of human participants in medical research [83]. The other 11 studies mentioned that they either had the consent of the relatives of the people with autism or their work was approved by the ethics committee of the given universities or other institutions. All details are provided in Table A.8.

Upon analyzing the obtained results, the majority of studies that indicated privacy and security aspects had obtained consent from family members, or the research was approved by ethical committees of universities/institutions; very few studies adhered to the Declaration of Helsinki. However, it is evident that there is insufficient consideration of privacy and security aspects in the majority of studies. Studies lacking pertinent details may not adhere to ethical protocols, thereby generating concerns about the protection of participants.

**4. Discussion**

*4.1. Findings*

The great majority of the studies analyzed referred to the autism spectrum disorder in different ways. Namely, it was noticed the use of two terminologies "autism" and "all kinds of autism" to refer to this condition interchangeably. This shows a lack of unification on the use of this terminology by the scientific community of the HER field. More importantly, more research needs to be placed towards mild and high-functioning autism, as well as other conditions of the autism spectrum like Asperger, which according to the results are just marginally considered.

While a majority of studies indicate the type of autism considered in their research, a significant number did not. Omitting information on the type of autism considered in studies can hinder accurate interpretation and reduce the applicability of findings, leading to potential misinterpretations and limiting the generalizability of research outcomes. Additionally, the absence of this specification may impede meaningful comparisons across studies, hindering the overall advancement of knowledge in the field of emotion recognition in autism.

The lack of proper specification of gender aspects in over half of the reviewed studies can have several consequences. It may lead to an incomplete understanding of how gender influences the outcomes of the research, potentially masking gender-related patterns or differences in emotional recognition within the context of ASD. Additionally, it hinders the generalizability of findings, as the impact of gender on emotion recognition might be relevant.

The number of participants involved in the studies varies remarkably, thus limiting the comparability of the results. While it is generally encouraged in the area to include as many participants as possible, the number of involved individuals should be fairly supported via an appropriate statistical power analysis. At least, it should be attempted to guarantee a sufficient number of participants matching the average number of the art.

The emotion recognition modality most predominantly used is the one based on facial expressions, followed by speech. The reason for favouring the measurement of physical variables over physiological might be related to the fact that emotions are socially expressed and perceived via physical cues, such as facial and visual expressions and the voice tone. ASD is however sometimes characterized by a lack of expressiveness. Hence, it might be a good choice to observe and to analyze physiological behaviour in this population in addition to the physical one.

A major part of the studies used the six basic universal emotions (anger, sadness, happiness, disgust, surprise, and fear) considered as a standard for HER systems. The reasons may have to do with the fact that such emotions represent the most common set expressed by people in their daily life [82]. Moreover, using emotions similar to the ones used in prior works facilitate cross comparison and reproducibility, so that better conclusions can be drawn. Several studies used combinations of the six basic emotions by adding very specific emotions (neutral, calm, nervous, scared, curious, excited, sleepy, contempt, joy, and contentment). From this set of emotions, "neutral" stands out as the most frequent one, possibly due to its prevalence in the daily life.

The majority of devices and sensors employed in the period 2015-2023 are seen to be particularly advances with respect to the ones used in the period 2011-2014. IP and infrared cameras, face or body tracking sensors, and partially wearable sensors or robots are used in the second half of the decade while more traditional systems such as webcams and microphones were used during the first half. From our analysis we can conclude that most works use a single technology to assess emotions in autism. The main reason for considering a sole device could be to simplify the sensor setup and lessen the intrusiveness sometimes felt by users when using these technologies. Combining multiple technologies to assess emotions may potentially lead to more accurate and robust decisions, as shown in the literature for neurotypical populations [84]. However, as we found out in a former study of ours [85], people with autism (adolescents) show reluctance to using multiple devices, and in particular to some specific ones such as infrared cameras.

All the algorithms used are of the supervised kind, which was expected since most of the reviewed studies are aimed at diagnosing autism. A limitation observed for such approaches is that they tend to be learned on general-purpose emotion recognition datasets, most likely due to a clear lack of existing autism-specific datasets. General-purpose datasets could serve well for boosting some machine learning models, however they are of limited use when it comes to recognising the emotions expressed by people with autism. One goal for the community could then be the creation of new relevant datasets particularly devised for autism applications. Moreover, we did not find any study exploring the use of unsupervised methods. The use of these methods allows for the creation of clusters, which could help identify people with similar patterns within a similar spectrum. This is found of much interest specially when it comes to a disorder like autism, which is quite diverse per se.

The performance results have been shown to vary among studies. High performances are obtained for a number of studies, however, it is observed that most of such studies do not describe in sufficient detail the evaluation method, thus hindering the validity of the reported results. Cross-validation and accuracy metrics are most widely used for evaluating the emotion recognition models performance [86]. A minority of the studies characterize the performance of their system more comprehensively using other metrics such as sensitivity and specificity. In order to avoid the effects of data bias, future research in this area is encouraged to consider using more robust metrics such as the F-score [87]. The number of data samples is also found key to determine the relevance of the reported results, and according to this review, only a minority of the studies appear to use a relevant sample set. This limits somewhat the validity of some of the results reported in the reviewed studies.

Privacy and security aspects have been partially addressed and only by a minority of the studies. It should be noted that this kind of studies work with sensitive data, and it is imperative to guarantee the protection of participants in medical research, especially when it comes to people with neurodevelopmental disorders. Presumably, the studies that did not give details on this information may not have followed any ethical protocol or perhaps simply forgot to report it in the manuscript. While the former is a more serious issue than the latter, we think this is an aspect that must be improved considerably in the future and information privacy and security should be compulsorily addressed in all studies of this nature.

### 4.2. Challenges and opportunities

From the previous findings a number of important challenges and opportunities are identified which should be considered in our opinion while designing, developing, and using emotion recognition technologies for people with autism.

One such challenge has to do with the heterogeneity of the autism spectrum disorder. The fact that autism is a spectrum disorder entails that people with autism have a wide range of abilities, difficulties, and preferences. Regular emotion recognition technologies may not account for this heterogeneity, as they often rely on standardized models and assumptions about emotional expressions of neurotypical persons. Hence, we consider important to build systems devised for each specific subtype of the disorder, and whether possible, favour the development of personalized approaches that consider the unique characteristics of each person with autism. A way to materialize this idea could be to use transfer learning approaches, in which an existing emotion recognition model trained on a large dataset from a heterogeneous cohort of individuals is used to learn a personalized model by tuning the former with new data of a particular individual or group of subjects pertaining to a specific autism subtype.

Another important challenge relates to the nonverbal communication variability linked to the disorder. People with autism may have atypical patterns in facial expressions, body language, and vocal tone. This review underscores that current emotion recognition technologies, predominantly reliant on visual and auditory cues, may struggle to accurately interpret these atypical communication styles. To overcome this, the creation of new algorithms tailored to the specific nonverbal communication of individuals with autism is proposed, potentially involving the development of expert-validated datasets that capture this variability [88,89]. While engaging individuals with autism in this process is ideal, an alternative involves using actors to mimic these cues based on expert instructions.

Many people with autism have sensory sensitivities, which can affect their tolerance for certain stimuli, such as bright lights, loud sounds, or touch. As a result, the use of emotion recognition technologies that involve a sensory input, like vivid displays, loudspeakers, or tactile sensors, may potentially cause discomfort or distress for some individuals with autism. Hence, it is important to make sure that the technologies are adaptable to the sensory needs and particularly to the preferences of the user. Emotion recognition systems should be designed to be flexible and compatible enough to operate with the sensor modalities chosen by the user. One way to achieve this is to develop ensemble learning models combining multiple individual models each running on data from different sensor modalities. By following this approach, the resulting emotion recognition model can easily adapt to the absence of one or various sensor modalities and still operate, although the accuracy of the system would be normally lower.

Other relevant challenge emphasizes the contextual nature of emotions, particularly pertinent in the case of autism where social nuances may pose difficulties for solely facial or vocal emotion recognition systems. To address this, integrating alternate sensing options, such as physiological cues, is suggested. However, the broader context, including situational cues, personal history, and individual preferences, is crucial for enhancing recognition accuracy. To capture this intricate information, incorporating virtual agents or chatbots in regular interactions with individuals with autism is proposed as a means to gather comprehensive data for more effective emotion recognition models.

As previously highlighted, emotion recognition technologies introduce significant ethical considerations concerning privacy and data

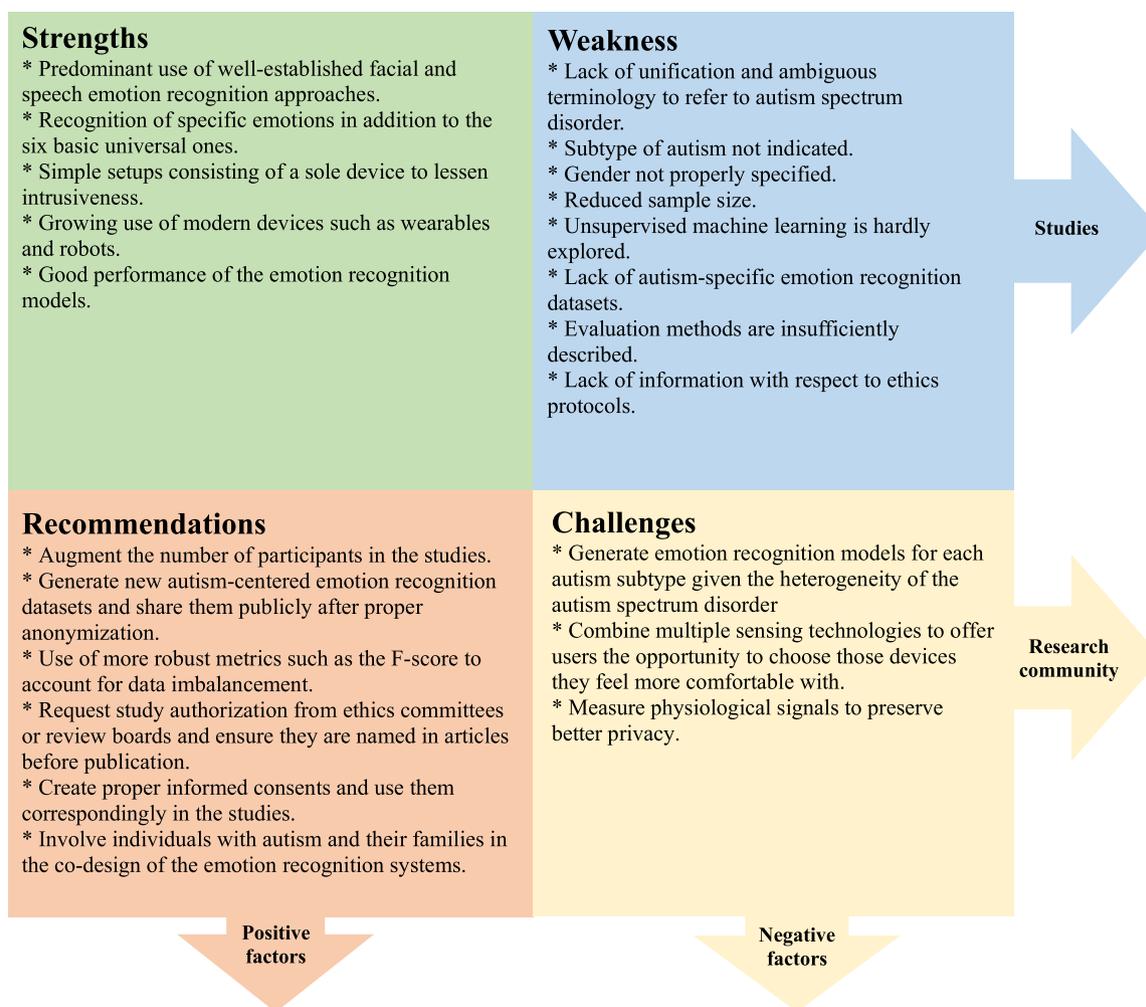

**Fig. 9.** Summary of the principal takeaways of the reviewed manuscripts.

security. The collection and storage of sensitive emotional data carry potential implications for an individual's privacy and autonomy. Therefore, it is paramount to establish rigorous privacy measures, secure informed consent, and guarantee that people with autism retain control over their personal information [90]. Given that many emotion recognition models operate on sensitive data, including video and audio, it becomes particularly imperative to transparently communicate the purpose, methodology, and safeguards associated with data collection. Ensuring these systems adhere to and assure full compliance with relevant regulations is essential. Consequently, the incorporation of emotion recognition technologies utilizing sensory inputs, such as vivid visual displays, loudspeakers, or tactile sensors, has the potential to induce discomfort or distress for some individuals with autism

Addressing these challenges requires interdisciplinary collaboration between researchers, technologists, and the autism community. It is crucial to involve individuals with autism and their families in the design and development of these technologies to ensure that they are respectful, inclusive, and beneficial for the target population.

A summary of the principal findings described previously is provided in Fig. 9. The diagram shows the strengths and weaknesses of the reviewed studies on sensing technologies and machine learning methods for emotion recognition in autism as well as the challenges and recommendations for the research community. Strengths are the aspects that these studies have performed well on and could be reproduced in future investigations. Weaknesses are matters that went wrong in these studies and could be improved in future research. Challenges are the elements that the scientific community needs to address successfully to boost the investigation of this topic. Recommendations are the suggestions for the research community working on this field.

*4.3. Limitations*

As for any other review, and despite having used a rather broad search strategy, it is certainly possible that some interesting studies may have left out from our analysis. Namely, the search areas of this systematic review were circumscribed to computer science and engineering respectively, as they are quite large areas and most relevant for the scope of this study. Nonetheless, it is also possible that some relevant studies indexed in other related categories may have been filtered out. We conducted a preliminary check for other domains such as psychology, behavioural sciences, or pediatrics and we did not find relevant studies that would meet the defined criteria. Another possible limitation of this review refers to the reference management software used to process both records and reports. We decided to use Mendeley since all reviewers were quite familiarised with the tool and all three contributing institutions supports access to it. One of the major advantages in deciding to use Mendeley is that it allows the creation of academic research communities through collaborative research [91]. However, other free and open-source reference management software, such as Zotero, could be more appropriate when it comes to pursuing open science principles. Finally, the protocol systematic review conducted here could have been pre-registered, for example via PROSPERO, however the researchers were not aware of this option when the work started. Nonetheless, we recently searched for similar pre-registered protocols and none resulted from the search, so we presume that no overlap exists between our work and other on-going reviews in the field.

## 5. Conclusion

Automatic emotion recognition constitutes a fairly consolidated research domain in the affective computing field. However, as it is shown in this review, its application to autism is limited and insufficiently validated. Thus, for example, new research should explore the design and development of models that account for the particular characteristics of people with autism, rather than pushing to the limits the generalisation of existing models trained on data collected from neurotypical people. In this regard, collecting and sharing publicly new datasets involving people with autism is found of paramount importance as these are practically nonexistent to date. More efforts should also be put towards describing in greater detail the characteristics of the samples subject to study. Gender, age, and autism type are not consistently reported thus making it difficult to assess the relevance of the proposed models and hindering the replicability of the studies. In view of the diverse nature of the autism spectrum, it seems also quite reasonable to explore in future studies the use of holistic sensing approaches. Indeed, facial expression recognition is ahead of other solutions, also in this domain, however, the disparity and lack of expressiveness among people with autism make it necessary to consider measuring multiple physical and physiological signals. Accomplishing these challenges demands interdisciplinary collaboration teams and the appropriate funding of governments and institutions to design, develop, and validate the required technologies from an autism-centric perspective and in realistic settings. We truly hope that reflecting on the positive contributions made by researchers in this field particularly on the ample room for improvement can spark great interest from other colleagues from the affective computing field to devote time and effort to boost this important domain.

## 6. Summary table

- Automatic emotion recognition constitutes a consolidated research domain in the affective computing field. Nevertheless, as shown in this review, its application to autism is limited and not sufficiently validated.
- New research should explore the design and development of models that take into account the unique characteristics of individuals with autism, rather than generalising existing models trained on data collected from neurotypical individuals.
- The collection and public sharing of new datasets that include individuals with autism is therefore considered of utmost importance, as they are virtually non-existent to date and should include more detailed characteristics of the samples under study.
- Considering the diverse nature of the autistic spectrum, it also seems quite reasonable to explore the use of holistic detection approaches in future studies.
- Facial expression recognition is ahead of other solutions, also in this area; nevertheless, the disparity and sometimes lack of expressiveness among individuals with autism makes it necessary to consider the measurement of multiple physical and physiological signals.
- Meeting these challenges requires interdisciplinary collaborative teams and adequate funding from governments and institutions to design, develop and validate the necessary technologies from an autism-centred perspective and in realistic settings.
- The overall aim is that reflection on the positive contributions made by researchers in this field, in particular on the vast room for improvement, may inspire great interest in other colleagues in the field of affective computing to dedicate time and effort to furthering this important field.

## CRediT authorship contribution statement

**Oresti Banos:** Writing – review & editing, Writing – original draft, Supervision, Methodology, Funding acquisition, Formal analysis, Conceptualization. **Zhoe Comas-González:** Visualization, Methodology, Formal analysis, Data curation. **Javier Medina:** Writing – review & editing, Writing – original draft, Visualization, Methodology, Funding acquisition, Formal analysis, Conceptualization. **Aurora Polo-Rodríguez:** Writing – original draft, Methodology, Investigation, Data curation. **David Gil:** Writing – review & editing, Writing – original draft, Methodology, Funding acquisition, Formal analysis, Conceptualization. **Jesús Peral:** Writing – review & editing, Writing – original draft, Supervision, Methodology, Funding acquisition, Formal analysis, Conceptualization. **Sandra Amador:** Writing – original draft, Visualization, Methodology, Data curation. **Claudia Villalonga:** Writing – original draft, Methodology, Investigation, Formal analysis, Data curation, Conceptualization.

## Declaration of competing interest

The authors declare that they have no known competing financial interests or personal relationships that could have appeared to influence the work reported in this paper.

## Acknowledgements

This research has been partially funded by the Spanish project "Advanced Computing Architectures and Machine Learning-Based Solutions for Complex Problems in Bioinformatics, Biotechnology, and Biomedicine (RTI2018-101674-B-I00)" and the Andalusian project "Integration of heterogeneous biomedical information sources by means of high performance computing. Application to personalized and precision medicine (P20_00163)". Funding for this research is provided by the EU Horizon 2020 Pharaon project 'Pilots for Healthy and Active Ageing' (no. 857188). Moreover, this research has received funding under the REMIND project Marie Sklodowska-Curie EU Framework for Research and Innovation Horizon 2020 (no. 734355). This research has been partially funded by the BALLADEER project (PROMETEO/2021/088) from the Conselleria de Innovación, Universidades, Ciencia y Sociedad Digital, Generalitat Valenciana. Furthermore, it has been partially funded by the AETHER-UA (PID2020-112540RB-C43) project from the Spanish Ministry of Science and Innovation. This work has been also partially funded by "La Conselleria de Innovación, Universidades, Ciencia y Sociedad Digital", under the project "Development of an architecture based on machine learning and data mining techniques for the prediction of indicators in the diagnosis and intervention of autism spectrum disorder. AICO/2020/117". This study was also funded by the Colombian Government through Minciencias grant number 860 "international studies for doctorate". This research has been partially funded by the Spanish Government by the project PID2021-127275OB-I00, FEDER "Una manera de hacer Europa". Moreover, this contribution has been supported by the Spanish Institute of Health ISCIII through the DTS21-00047 project. Furthermore, this work was funded by COST Actions "HARMONISATION" (CA20122) and "A Comprehensive Network Against Brain Cancer" (Net4Brain - CA22103). Sandra Amador is granted by the Generalitat Valenciana and the European Social Fund (CIACIF/ 2022/233).

## Appendix A. Tables

The tables referenced in the manuscript are included as appendices.

**Table A.1**
Main research goals for each reviewed study.

| Manuscript | Research goals |
| --- | --- |
| Piana et al. (2016) [48] | Develop a system for the automatic emotion recognition of emotions at realtime from the analysis of body movements exerted during serious gaming |
| Irani et al. (2018) [8] | Create a game to help children with autism cope with their emotional difficulties |
| Leo et al. (2015) [49] | Integrate automatic emotion recognition capabilities in a robot-children interaction tool for autism treatment |
| Postawka et al. (2019) [17] | Develop emotion recognition methods for behaviour model estimation based on body position |
| Jiang et al. (2018) [18] | Identify subjects with/without autism by using facial emotion recognition and eye tracking |
| Gao et al. (2015) [19] | Classify emotions through electroencephalography signals |
| Heni et al. (2016) [35] | Design an app to recognize both emotions and voice |
| Jeon et al. (2015) [50] | Examine how children with autism and neurotypical children understand and interpret emotions |
| Fan et al. (2017) [46] | Explore the feasibility of using electroencephalography signals to analyze the facial affect recognition process of individuals with autism |
| Joseph et al. (2018) [20] | Propose a new algorithm to detect primary emotions of children with autism in real time using deep learning |
| Spicker et al. (2016) [79] | Investigate the differences in perception and categorization of emotional facial expressions of virtual characters between children and adolescents with autism, attention-deficit hyperactivity disorder, and neurotypical ones |
| Enticott et al. (2014) [61] | Examine facial emotion recognition of matched static and dynamic images among adolescents with autism and adults and neurotypical individuals |
| Santhoshkumar et al. (2019) [21] | Predict basic emotions from children with autism using body movements |
| Sivasangari et al. (2019) [22] | Propose a new methods for the automatic recognition of emotions |
| Tang et al. (2017) [51] | Compare the manual tagging of emotions by teachers/parents with the automatic one produced by an automatic system during naturalistic tasks |
| Ley et al. (2019) [62] | Evaluate existing tools for emotion recognition based on facial features as well as vocal features in voice interactions |
| Chung et al. (2019) [52] | Develop an augmented reality system for the presentation of the emotions detected via a facial expression recognition model |
| Smitha et al. (2015) [53] | Determine a feasible method for realizing a portable emotion detector for children with autism |
| Daniels et al. (2018) [54] | Build a therapeutic tool for children with autism using wearable technologies to recognize emotions as well as estimate how these interpretations differ between children with autism and neurotypical children |
| Smitha et al. (2013) [55] | Build a hardware efficient portable emotion recognizer on an FPGA to aid children with autism during the recognition of emotions |
| Tang et al. (2016) [56] | Develop an IoT natural play environment to help neurotypical children to understand children with autism emotions |
| Liliana et al. (2020) [23] | Develop an artificial intelligent model based on psychological knowledge to recognize emotions by analyzing facial expressions |
| Ghorbandaei et al. (2018) [72] | Build a robotic platform for reciprocal interaction in which a vision system recognizes the facial expressions of the user through a fuzzy clustering method |
| Elamir et al. (2018) [24] | Design an automatic emotion recognition system based on nonlinear analysis of various physiological signals |
| Fernandes et al. (2011) [36] | Apply a game-based approach to teach children with autism to recognize facial emotions using realtime automatic facial expression analysis and virtual character synthesis |
| Anishchenko et al. (2017) [37] | Develop a tablet application for learning and detecting facial expressions |
| Su et al. (2018) [78] | Examine the differences of emotion recognition and eye gaze pattern between children with autism and neurotypical ones using facial expressions |
| Arellano et al. (2015) [75] | Assess how abstract emotional facial expressions influence the categorization of the emotions by children and adolescents with high functioning autism |
| AndleebSiddiqui et al. (2020) [25] | Recognize emotions via speech analysis using deep learning |
| Globerson et al. (2012) [77] | Explore the association between psychoacoustic abilities and vocal emotion recognition in a group of individuals with autism and a matched group of neurotypical individuals |
| Sunitha et al. (2014) [73] | Collect a new dataset for emotion recognition from speech |
| Bagirathan et al. (2020) [47] | Compare psycho-physiological signals from neurotypical children and children with autism |
| Piparsaniyan et al. (2014) [26] | Propose a new method for facial expression recognition |
| Marchi et al. (2012) [27] | Classify a number of emotions in different scenarios |
| Marchi et al. (2015) [74] | Analyze various existing emotion recognition datasets |
| Syeda et al. (2017) [57] | Perform visual face scanning pattern and emotion perception analysis between neurotypical children and children with autism |
| Guha et al. (2018) [76] | Assess the reduced complexity in facial expression dynamics of subjects with high functional autism relative to their neurotypical peers |
| Costescu et al. (2020) [58] | Test the effectiveness of a facial expression recognition instrument in both neurotypical individuals and adolescents with autism |
| Tracy et al. (2011) [80] | Show impaired recognition of all basic emotion expressions and more socially complex ones when forced to complete the recognition process in a very brief time frame |
| Chung et al. (2020) [52] | Design an e-learning model for students with autism |
| Zhang et al. (2016) [60] | Propose a new emotion recognition system based on facial expression images |
| Dantas et al. (2022) [38] | Build a game to support the ability for children with autism to recognize and express basic emotions |
| Saranya et al. (2022) [28] | Develop an deep learning-based emotion recognition method for improving the rate of detection in children with autism |
| Sukumaran et al. (2021) [63] | Identify the presence of ASD and to analyze the emotions of children with autism through their voices |
| Wang et al. (2021) [30] | Analyze an emotion care system based on big data analysis for autism disorder patient training, where emotion is detected in terms of facial expression |
| Banire et al. (2021) [29] | Develop a face-based attention recognition model using geometric feature transformation and time-domain spatial features |
| Piana et al. (2021) [39] | Build a system for the automatic emotion recognition designed for helping children with autism to learn to recognize and express emotions by means of their full-body movement |
| Ruan et al. (2022) [64] | Design and build automatic computer-based learning tools for children with ASD to improve their performance in Maths |
| Milling et al. (2022) [65] | Contribute with a voice activity detection (VAD) system specifically adapted to children with autism vocalisations |
| Chitre et al. (2022) [66] | Model a Real-time Speech Emotion Recognition (SER) that takes audio signals as inputs and detects the emotions based on those signals |
| Wang et al. (2022) [67] | Examine the effects of video-based intervention on emotion recognition in four children with ASD with imitation in speech |
| Wan et al. (2022) [40] | Propose a novel framework for human-computer human-robot interaction and introduce a preliminary intervention study for improving the emotion recognition of Chinese children with autism |
| Silva et al. (2021) [41] | Develop a system capable of automatically detecting emotions through facial expressions and interfacing them with a robotic platform to allow social interaction with children with ASD |
| Praveena et al. (2021) [68] | Recognize and predict face emotion in ASD |
| Rojas et al. (2021) [42] | Help people with a degree of difficulty in interpreting emotions so that they can have a normal social interaction through a mobile app in real-time |
| Karanchery et al. (2021) [43] | Provide a solution to be deployed in learning environments for individuals with ASD to aid the primary caregivers in understanding their emotional states |
| Valles et al. (2021) [69] | Develop a speech emotion recognition system to help children with autism to better identify the emotions of their communication partner |
| DIzicheh et al. (2021) [44] | Present a serious game called EmoAnim that utilizes animations to screen players' emotion recognition capabilities |

**Table A.1** (*continued*)

| Manuscript | Research goals |
| --- | --- |
| Pulido-Castro et al. (2021) [31] | Develop a real-time emotional recognition algorithm based on facial expressions |
| Arabian et al. (2021) [32] | Highlight the significance of image pre-processing in Deep Neural Network models for facial expression recognition to improve training, overall accuracy and efficacy |
| Li et al. (2021) [33] | Introduce a novel way to combine human expertise and machine intelligence for ASD affect recognition via a two-stage schema |
| Ghanouni et al. (2021) [45] | Develop a novel motion game to address perspective by incorporating both children/youth with ASD and their parents feedback |
| Zhang et al. (2023) [70] | Develop a novel discriminative few shot learning method to analyze hour-long video data and explore the fusion of facial dynamics for automatic ASD trait classification |
| Talaat (2023) [34] | Develop real-time emotion recognition system based on deep learning neural networks for youngsters with autism |
| Murugaiyan et al. (2023) [71] | Propose a model to help people with ASD to understand other's sentiments expressed through speech |

**Table A.2**
Autism types investigated in each reviewed study.

| Manuscript | Autism type (or term used) |
| --- | --- |
| Piana et al. (2016) [48] | Autism |
| Irani et al. (2018) [8] | Autism |
| Leo et al. (2015) [49] | Autism |
| Postawka et al. (2019) [17] | Autism |
| Jiang et al. (2019) [18] | Autism |
| Gao et al. (2015) [19] | Autism |
| Heni et al. (2016) [35] | Autism |
| Jeon et al. (2015) [50] | Autism |
| Fan et al. (2017) [46] | Autism |
| Joseph et al. (2018) [20] | Autism |
| Spicker et al. (2016) [79] | Attention deficit hyperactivity disorder |
| Enticott et al. (2014) [61] | Autism |
| Santhoshkumar et al. (2019) [21] | Autism |
| Sivasangari et al. (2019) [22] | Autism |
| Tang et al. (2017) [51] | Autism |
| Ley et al. (2019) [62] | Autism |
| Chung et al. (2019) [52] | Autism |
| Smitha et al. (2015) [53] | Autism |
| Daniels et al. (2018) [54] | Autism |
| Smitha et al. (2013) [55] | Autism |
| Tang et al. (2016) [56] | Autism |
| Liliana et al. (2020) [23] | Autism |
| Ghorbandaei et al. (2018) [72] | All kind of autism |
| Elamir et al. (2018) [24] | All kind of autism |
| Fernandes et al. (2011) [36] | All kind of autism, Asperger |
| Anishchenko et al. (2017) [37] | All kind of autism |
| Su et al. (2018) [78] | Mild autism |
| Arellano et al. (2015) [75] | High-functioning autism |
| AndleebSiddiqui et al. (2020) [25] | All kind of autism |
| Globerson et al. (2012) [77] | High-functioning autism |
| Sunitha et al. (2014) [73] | All kind of autism |
| Bagirathan et al. (2020) [47] | Middle and moderate autism |
| Piparsaniyan et al. (2014) [26] | All kind of autism |
| Marchi et al. (2012) [27] | All kind of autism |
| Marchi et al. (2015) [74] | All kind of autism |
| Syeda et al. (2017) [57] | Autism |
| Guha et al. (2018) [76] | High-functioning autism |
| Costescu et al. (2020) [58] | Autism |
| Tracy et al. (2011) [80] | High-functioning autism, Asperger |
| Chung et al. (2020) [52] | Autism |
| Zhang et al. (2016) [60] | Autism |
| Dantas et al. (2022) [38] | Autism |
| Saranya et al. (2022) [28] | Autism |
| Sukumaran et al. (2021) [63] | Autism |
| Wang et al. (2021) [30] | Autism |
| Banire et al. (2021) [29] | Autism |
| Piana et al. (2021) [39] | Autism |
| Ruan et al. (2022) [64] | Autism |
| Milling et al. (2022) [65] | Autism |
| Chitre et al. (2022) [66] | Autism |
| Wang et al. (2022) [67] | Autism |
| Wan et al. (2022) [40] | Autism |
| Silva et al. (2021) [41] | High-functioning autism, Asperger |
| Praveena et al. (2021) [68] | Autism |
| Rojas et al. (2021) [42] | Autism |
| Karanchery et al. (2021) [43] | Not mentioned |
| Valles et al. (2021) [69] | Autism |
| DIzicheh et al. (2021) [44] | Autism |
| Pulido-Castro et al. (2021) [31] | Autism |
| Arabian et al. (2021) [32] | Autism |
| Li et al. (2021) [33] | Autism |
| Ghanouni et al. (2021) [45] | High-functioning autism |
| Zhang et al. (2023) [70] | Autism |
| Talaat (2023) [34] | Autism |
| Murugaiyan et al. (2023) [71] | Autism |

**Table A.3**
Emotional expressions and sensed body regions for each reviewed study.

| Manuscript | Emotion recognition modality | Sensed body region |
| --- | --- | --- |
| Piana et al. (2016) [48] | Body movement emotion recognition | Body movement |
| Irani et al. (2018) [8] | Facial emotion recognition | Facial expression |
| Leo et al. (2015) [49] | Facial emotion recognition | Facial expression |
| Postawka et al. (2019) [17] | Emotion recognition based on activities | Not sufficiently described |
| Jiang et al. (2019) [18] | Multimodal emotion recognition | Facial expression, Eye tracking |
| Gao et al. (2015) [19] | Brain activity emotion recognition | EEG |
| Heni et al. (2016) [35] | Facial emotion recognition | Facial expression, Voice expression |
| Jeon et al. (2015) [50] | Multimodal emotion recognition | Facial expression, Voice expression |
| Fan et al. (2017) [46] | Facial emotion recognition | Facial expression |
| Joseph et al. (2018) [20] | Facial emotion recognition | Facial expression |
| Spicker et al. (2016) [79] | Facial emotion recognition | Facial expression |
| Enticott et al. (2014) [61] | Facial emotion recognition | Facial expression |
| Santhoshkumar et al. (2019) [21] | Body emotion recognition | Body movement |
| Sivasangari et al. (2019) [22] | Facial emotion recognition | Facial expression |
| Tang et al. (2017) [51] | Facial emotion recognition | Facial expression |
| Ley et al. (2019) [62] | Multimodal emotion recognition | Facial expression, Voice expression |
| Chung et al. (2019) [52] | Facial emotion recognition | Facial expression |
| Smitha et al. (2015) [53] | Facial emotion recognition | Facial expression |
| Daniels et al. (2018) [54] | Facial emotion recognition | Facial expression |
| Smitha et al. (2013) [55] | Facial emotion recognition | Facial expression |
| Tang et al. (2016) [56] | Facial emotion recognition | Facial expression |
| Liliana et al. (2020) [23] | Facial emotion recognition | Facial expression |
| Ghorbandaei et al. (2018) [72] | Facial emotion recognition | Facial expression |
| Elamir et al. (2018) [24] | Multimodal emotion recognition | EEG, EMG |
| Fernandes et al. (2011) [36] | Facial emotion recognition | Facial expression |
| Anishchenko et al. (2017) [37] | Facial emotion recognition | Facial expression |
| Su et al. (2018) [78] | Visual emotion recognition | Eye tracking |
| Arellano et al. (2015) [75] | Facial emotion recognition | Facial expression |
| AndleebSiddiqui et al. (2020) [25] | Speech emotion recognition | Voice expression |
| Globerson et al. (2012) [77] | Speech emotion recognition | Voice expression |
| Sunitha et al. (2014) [73] | Speech emotion recognition | Voice expression |
| Bagirathan et al. (2020) [47] | Multimodal emotion recognition | EEG, EMG, HR |
| Piparsaniyan et al. (2014) [26] | Facial emotion recognition | Facial expression |
| Marchi et al. (2012) [27] | Speech emotion recognition | Voice expression |
| Marchi et al. (2015) [74] | Speech emotion recognition | Voice expression |
| Syeda et al. (2017) [57] | Multimodal emotion recognition | Facial expression, Eye tracking |
| Guha et al. (2018) [76] | Facial emotion recognition | Facial expression |
| Costescu et al. (2020) [58] | Facial emotion recognition | Facial expression |
| Tracy et al. (2011) [80] | Facial emotion recognition | Facial expression |
| Chung et al. (2020) [52] | Facial emotion recognition | Facial expression |
| Zhang et al. (2016) [60] | Facial emotion recognition | Facial expression |
| Dantas et al. (2022) [38] | Facial emotion recognition | Facial expression |
| Saranya et al. (2022) [28] | Facial emotion recognition | Facial expression |
| Sukumaran et al. (2021) [63] | Speech emotion recognition | Voice expression |
| Wang et al. (2021) [30] | Facial emotion recognition | Facial expression |
| Banire et al. (2021) [29] | Facial emotion recognition | Facial expression |
| Piana et al. (2021) [39] | Body movement emotion recognition | Body movement |
| Ruan et al. (2022) [64] | Facial emotion recognition | Facial expression |
| Milling et al. (2022) [65] | Speech emotion recognition | Voice expression |
| Chitre et al. (2022) [66] | Speech emotion recognition | Voice expression |
| Wang et al. (2022) [67] | Facial emotion recognition | Facial expression |
| Wan et al. (2022) [40] | Facial emotion recognition | Facial expression |
| Silva et al. (2021) [41] | Facial emotion recognition | Facial expression |
| Praveena et al. (2021) [68] | Facial emotion recognition | Facial expression |
| Rojas et al. (2021) [42] | Facial emotion recognition | Facial expression |
| Karanchery et al. (2021) [43] | Facial emotion recognition | Facial expression |
| Valles et al. (2021) [69] | Speech emotion recognition | Voice expression |
| DIzicheh et al. (2021) [44] | Facial emotion recognition | Facial expression |
| Pulido-Castro et al. (2021) [31] | Facial emotion recognition | Facial expression |
| Arabian et al. (2021) [32] | Facial emotion recognition | Facial expression |
| Li et al. (2021) [33] | Multimodal emotion recognition | Facial expression, Voice expression |
| Ghanouni et al. (2021) [45] | Facial emotion recognition | Facial expression |
| Zhang et al. (2023) [70] | Multimodal emotion recognition | Facial expression, Voice expression, Eye tracking, Body movement |
| Talaat (2023) [34] | Facial emotion recognition | Facial expression |
| Murugaiyan et al. (2023) [71] | Speech emotion recognition | Voice expression |

**Table A.4**
Sample characteristics for each reviewed study.

| Manuscript | Sample size | Sample type |
| --- | --- | --- |
| Piana et al. (2016) [48] | 60 | Disorder undefined. Gender undefined. Age undefined |
| Irani et al. (2018) [8] | 68 | 23 children with ASD, 6 children with low ASD, 17 children with high ASD. 22 children without autism. Gender undefined. 6-14 years old |
| Leo et al. (2015) [49] | Not sufficiently described | External dataset |
| Postawka et al. (2019) [17] | Not sufficiently described | Not sufficiently described |
| Jiang et al. (2019) [18] | 58 | 23 people with ASD, 35 people without autism. 13 females, 45 males. 8-34 years old |
| Gao et al. (2015) [19] | 21 | Not sufficiently described |
| Heni et al. (2016) [35] | Not sufficiently described | Not sufficiently described |
| Jeon et al. (2015) [50] | 11 | 11 children without autism. 2 females, 9 males. Age undefined |
| Fan et al. (2017) [46] | 8 | 8 people with high ASD. 8 males. 13-18 years old |
| Joseph et al. (2018) [20] | Not sufficiently described | Not sufficiently described |
| Spicker et al. (2016) [79] | 62 | 16 people with high ASD, 24 people with attention-deficit hyperactivity disorder, 22 people without autism. 62 males. 13-18 years |
| Enticott et al. (2014) [61] | 36 | 36 people with ASD. Gender undefined. Age undefined |
| Santhoshkumar et al. (2019) [21] | 10 | 10 children with autism. Gender undefined. 5-11 years old |
| Sivasangari et al. (2019) [22] | 500 | 500 people with ASD. Gender undefined. Age undefined |
| Tang et al. (2017) [51] | 6 | 6 children with autism. Gender undefined. 4-5 years old |
| Ley et al. (2019) [62] | 21 | 21 people with ASD. Gender undefined. 23-39 years old |
| Chung et al. (2019) [52] | 6 | 6 people without autism. 6 female. 22 years old |
| Smitha et al. (2015) [53] | 10 | 10 people without autism. 10 females. Age undefined |
| Daniels et al. (2018) [54] | 43 | 23 people with autism, 20 people without autism. 10 females, 33 males. 11-12 years old |
| Smitha et al. (2013) [55] | 10 | 10 people without autism. 10 females. Age undefined |
| Tang et al. (2016) [56] | Not sufficiently described | Not sufficiently described |
| Liliana et al. (2020) [23] | 262 | Disorder undefined. 176 females, 86 males. 17-50 years old |
| Ghorbandaei et al. (2018) [72] | 14 | 14 children with autism. 4 females, 10 males. 4-5 years old |
| Elamir et al. (2018) [24] | 32 | Disorder undefined. 16 females, 16 males. 19-37 years old |
| Fernandes et al. (2011) [36] | 145 | Disorder undefined. 75 females, 70 males. 6-12 years old |
| Anishchenko et al. (2017) [37] | 19 | 19 children with autism. 2 females, 17 males. 6-12 years old |
| Su et al. (2018) [78] | 29 | 10 children with autism, 19 children without autism. Gender undefined. 5-7 years old |
| Arellano et al. (2015) [75] | 39 | 17 teenagers with autism, 22 teenagers without autism. Gender undefined. 14-18 years old |
| AndleebSiddiqui et al. (2020) [25] | 188 | 94 children with autism, 94 children without autism. Gender undefined. 10-13 years old |
| Globerson et al. (2012) [77] | 55 | 23 people with autism, 32 people without autism. 55 males. 20-39 years old |
| Sunitha et al. (2014) [73] | 25 | Disorder undefined. 12 females, 13 males. 5-12 years old |
| Bagirathan et al. (2020) [47] | 12 | 6 children with autism, 6 children without autism. Gender undefined. 7-11 years old |
| Piparsaniyan et al. (2014) [26] | 10 | Disorder undefined. 10 females. Age undefined |
| Marchi et al. (2012) [27] | 20 | 9 children with autism, 11 children without autism. 6 females, 14 males. 5-12 years old |
| Marchi et al. (2015) [74] | 56 | 25 children with autism, 31 children without autism. 21 females, 35 males. 5-11 years old |
| Syeda et al. (2017) [57] | 42 | 21 subjects with autism, 21 subjects without autism. 14 females, 28 males. 5-17 years old |
| Guha et al. (2018) [76] | 39 | 20 children with high ASD, 19 children without autism. 3 females, 36 males. 9-14 years old |
| Costescu et al. (2020) [58] | 51 | 11 children with autism, 40 children without autism. Gender undefined. 2-14 years old |
| Tracy et al. (2011) [80] | 28 | 11 children with high ASD, 15 children with Asperger, 2 children with PDDNOS. Gender undefined. 12 years old |
| Chung et al. (2020) [52] | Not sufficiently described | Not sufficiently described |
| Zhang et al. (2016) [60] | 20 | Disorder undefined. Gender undefined. 20-35 years old |
| Dantas et al. (2022) [38] | 8 | 4 children with autism, 4 children without autism. Gender undefined. 6-12 years old |
| Saranya et al. (2022) [28] | Not sufficiently described | Not sufficiently described |
| Sukumaran et al. (2021) [63] | Not sufficiently described | Not sufficiently described |
| Wang et al. (2021) [30] | 15 | 15 people without autism. 5 females, 10 males. 23 - 60 years old |
| Banire et al. (2021) [29] | 46 | 20 children with autism. 26 people without autism. 34 boys, 12 girls. 7 - 11 years old |
| Piana et al. (2021) [39] | 10 | 10 children with high functioning ASD. 1 girl, 9 boys. 8 - 11 years old |
| Ruan et al. (2022) [64] | Not sufficiently described | Not sufficiently described |
| Milling et al. (2022) [65] | 25 | 25 children with autism. 6 females, 19 males. 8 years old |
| Chitre et al. (2022) [66] | Not sufficiently described | Not sufficiently described |
| Wang et al. (2022) [67] | 4 | 4 children with autism. Gender undefined. 4 - 8 years old |
| Wan et al. (2022) [40] | 10 | 10 children with autism. Gender undefined. 3 - 8 years old |
| Silva et al. (2021) [41] | 37 | 6 children with high functioning ASD. 31 children without autism. Gender undefined. 6 - 9 years old |
| Praveena et al. (2021) [68] | Not sufficiently described | Not sufficiently described |
| Rojas et al. (2021) [42] | Not sufficiently described | Not sufficiently described |
| Karanchery et al. (2021) [43] | Not sufficiently described | Not sufficiently described |
| Valles et al. (2021) [69] | Not sufficiently described | Not sufficiently described |
| DIzicheh et al. (2021) [44] | 25 | 10 children with autism, 15 children without autism. Gender undefined. 7-8 years old |
| Pulido-Castro et al. (2021) [31] | Not sufficiently described | Not sufficiently described |
| Arabian et al. (2021) [32] | 80 | 80 people with autism. Gender undefined. Age undefined |
| Li et al. (2021) [33] | 6 | 6 children with autism. Gender undefined. Age undefined |
| Ghanouni et al. (2021) [45] | 20 | 4 children with high functioning ASD, 6 youth with high ASD, 10 adults without autism. Gender undefined. Age undefined |
| Zhang et al. (2023) [70] | 33 | 33 people with autism. 7 females, 26 males. 16 - 37 years old |
| Talaat (2023) [34] | Not sufficiently described | Not sufficiently described |
| Murugaiyan et al. (2023) [71] | Not sufficiently described | Not sufficiently described |

**Table A.5**
Emotions used for the training-validation and testing of the recognition models in the reviewed studies.

| Manuscript | Training and validation | Test |
| --- | --- | --- |
| Piana et al. (2016) [48] | Anger, sadness, happiness, disgust, surprise, fear | Anger, sadness, happiness, disgust, surprise, fear |
| Irani et al. (2018) [8] | Anger, sadness, happiness, fear | Anger, sadness, happiness, fear |
| Leo et al. (2015) [49] | Anger, sadness, happiness, disgust, surprise, fear | Anger, sadness, happiness, disgust, surprise, fear |
| Postawka et al. (2019) [17] | Calm, nervous, angry/aggressive | Calm, nervous, angry/aggressive |
| Jiang et al. (2019) [18] | Anger, sadness, happiness, disgust, surprise, fear | Anger, sadness, happiness, disgust, surprise, fear |
| Gao et al. (2015) [19] | Happy, calm, sad, scared | Happy, calm, sad, scared |
| Heni et al. (2016) [35] | Anger, sadness, happiness, disgust, surprise, fear | Anger, sadness, happiness, disgust, surprise, fear |
| Jeon et al. (2015) [50] | Curious, excited, happy, neutral, sad, scared, sleepy | Curious, excited, happy, neutral, sad, scared, sleepy |
| Fan et al. (2017) [46] | Anger, contempt, disgust, fear, joy, sadness, surprise, neutral | Anger, contempt, disgust, fear, joy, sadness, surprise, neutral |
| Joseph et al. (2018) [20] | Anger, sadness, happiness, disgust, surprise, fear | Anger, sadness, happiness, disgust, surprise, fear |
| Spicker et al. (2016) [79] | Anger, sadness, happiness, disgust, surprise, fear | Anger, sadness, happiness, disgust, surprise, fear |
| Enticott et al. (2014) [61] | Anger, sadness, happiness, disgust, surprise, fear | Anger, sadness, happiness, disgust, surprise, fear |
| Santhoshkumar et al. (2019) [21] | Happy, angry, sad, fear, neutral | Happy, angry, sad, fear, neutral |
| Sivasangari et al. (2019) [22] | Anger, sadness, happiness, disgust, surprise, fear | Anger, sadness, happiness, disgust, surprise, fear |
| Tang et al. (2017) [51] | Happiness, non-happiness | Happiness, non-happiness |
| Ley et al. (2019) [62] | Contentment, surprise, anger, sadness, disgust, fear, joy, happy, neutral | Contentment, surprise, anger, sadness, disgust, fear, joy, happy, neutral |
| Chung et al. (2019) [52] | Happiness, surprise, anger, sadness, fear, disgust, neutral | Happiness, surprise, anger, sadness, fear, disgust, neutral |
| Smitha et al. (2015) [53] | Happiness, surprise, anger, sadness, fear, disgust, neutral | Happiness, surprise, anger, sadness, fear, disgust, neutral |
| Daniels et al. (2018) [54] | Happiness, surprise, anger, sadness, fear, disgust, neutral | Happiness, surprise, anger, sadness, fear, disgust, neutral |
| Smitha et al. (2013) [55] | Happiness, surprise, anger, sadness, neutral | Happiness, surprise, anger, sadness, neutral |
| Tang et al. (2016) [56] | Anger, sadness, happiness, disgust, surprise, fear | Anger, sadness, happiness, disgust, surprise, fear |
| Liliana et al. (2020) [23] | Anger, sadness, happiness, disgust, surprise, fear | Anger, sadness, happiness, disgust, surprise, fear |
| Ghorbandaei et al. (2018) [72] | Anger, sadness, happiness, disgust, surprise, fear, neutral | Anger, sadness, happiness, disgust, surprise, fear, neutral |
| Elamir et al. (2018) [24] | Happy, neutral, sad, excited, calm, sleepy | Not sufficiently described |
| Fernandes et al. (2011) [36] | Not sufficiently described | Not sufficiently described |
| Anishchenko et al. (2017) [37] | Not sufficiently described | Anger, sadness, happiness, disgust, surprise, fear |
| Su et al. (2018) [78] | Not sufficiently described | Anger, sadness, happiness, disgust, surprise, fear |
| Arellano et al. (2015) [75] | Not sufficiently described | Anger, sadness, happiness, disgust, surprise, fear |
| AndleebSiddiqui et al. (2020) [25] | Angry, happy, neutral, sad | Anger, neutral, fear, happiness, sadness |
| Globerson et al. (2012) [77] | Neutral, happiness, sadness, anger, fear | Anger, sadness, happiness, disgust, surprise, fear |
| Sunitha et al. (2014) [73] | Anger, neutral, fear, happiness, sadness | Anger, sadness, happiness, disgust, surprise, fear |
| Bagirathan et al. (2020) [47] | Anger, neutral, fear, happiness, sadness | Anger, sadness, happiness, disgust, surprise, fear |
| Piparsaniyan et al. (2014) [26] | Anger, disgust, fear, happiness, sadness, surprise, neutral | Anger, sadness, happiness, disgust, surprise, fear |
| Marchi et al. (2012) [27] | Happy, sadness, angry, surprised, afraid, ashamed, calm, proud, neutral | Anger, sadness, happiness, disgust, surprise, fear |
| Marchi et al. (2015) [74] | Happy, sadness, angry, fearful, surprised, neutral | Anger, sadness, happiness, disgust, surprise, fear |
| Syeda et al. (2017) [57] | Anger, sadness, happiness, disgust, surprise, fear | Anger, sadness, happiness, disgust, surprise, fear |
| Guha et al. (2018) [76] | Happy, sadness, angry, fearful, surprised, neutral | Anger, sadness, happiness, disgust, surprise, fear |
| Costescu et al. (2020) [58] | Happy, sadness, angry, fear | Anger, sadness, happiness, disgust, surprise, fear |
| Tracy et al. (2011) [80] | Happy, sadness, angry, fear, surprised, pride | Anger, sadness, happiness, disgust, surprise, fear |
| Chung et al. (2020) [52] | Happy, sadness, angry, fearful, surprised | Anger, sadness, happiness, disgust, surprise, fear |
| Zhang et al. (2016) [60] | Happy, sadness, angry, fearful, surprised | Anger, sadness, happiness, disgust, surprise, fear |
| Dantas et al. (2022) [38] | Happy, sadness, anger, surprise, disgust, fear | Happy, sadness, anger, surprise, disgust, fear |
| Saranya et al. (2022) [28] | Not sufficiently described | Not sufficiently described |
| Sukumaran et al. (2021) [63] | Anger, disgust, neutral, happiness, calmness, fear, sadness | Anger, disgust, neutral, happiness, calmness, fear, sadness |
| Wang et al. (2021) [30] | Angry, disgust, fear, happy, sad, surprise, neutral | Six emotions (not specified) |
| Banire et al. (2021) [29] | Not sufficiently described | Not sufficiently described |
| Piana et al. (2021) [39] | Happiness, Sadness, Anger, Fear | Happiness, Sadness, Anger, Fear |
| Ruan et al. (2022) [64] | Contempt, Happiness, Fear, Neutrality, Disgust, Anger, Surprise, Sadness | Contempt, Happiness, Fear, Neutrality, Disgust, Anger, Surprise, Sadness |
| Milling et al. (2022) [65] | Arousal / Valence | Arousal / Valence |
| Chitre et al. (2022) [66] | Happy, sad, anger, fear, surprise, neutral and disgust | Happy, sad, anger, fear, surprise, neutral and disgust |
| Wang et al. (2022) [67] | Not sufficiently described | Happy, angry, afraid, sad, surprised, disgusted |
| Wan et al. (2022) [40] | Happiness, sadness, fear, anger | Happiness, sadness, fear, anger |
| Silva et al. (2021) [41] | Happiness, sadness, anger, surprise, fear | Happiness, sadness, anger, surprise, fear |
| Praveena et al. (2021) [68] | Happy, Sad, Fear, Neutral, Surprise, Angry | Happy, Sad, Fear, Neutral, Surprise, Angry |
| Rojas et al. (2021) [42] | Scared, Disgusted, Happy, Sad, Angry, Surprised, Contempt, Neutral | Scared, Disgusted, Happy, Sad, Angry, Surprised, Contempt, Neutral |
| Karanchery et al. (2021) [43] | Anger, Disgust, Fear, Happy, Sad, Surprise, Neutral | Anger, Disgust, Happiness, Neutral, Surprise |
| Valles et al. (2021) [69] | Happiness, Sadness, Surprise, Anger, Fear, Disgust | Not sufficiently described |
| DIzicheh et al. (2021) [44] | Not sufficiently described | Fear, Sadness, Happiness, Anger |
| Pulido-Castro et al. (2021) [31] | Anger, Disgust, Happiness, Neutral, Surprise | Anger, Disgust, Happiness, Neutral, Surprise |
| Arabian et al. (2021) [32] | Anger, Disgust, Fear, Happiness, Sadness, Surprise | Anger, Disgust, Fear, Happiness, Sadness, Surprise |
| Li et al. (2021) [33] | Neutral, Interested, Positive, Positive and talking, Odd positive, Negative | Neutral, Interested, Positive, Positive and talking, Odd positive, Negative |
| Ghanouni et al. (2021) [45] | Not sufficiently described | Anger, Disgust, Happiness, Neutral, Surprise |
| Zhang et al. (2023) [70] | Not sufficiently described | Not sufficiently described |
| Talaat (2023) [34] | Surprise, Delight, Sadness, Fear, Joy, Anger | Surprise, Delight, Sadness, Fear, Joy, Anger |
| Murugaiyan et al. (2023) [71] | Happy, Sad, Angry, Calm, Fear, Neutral, Disgust, Surprise, Boredom | Happy, Sad, Angry, Calm, Fear, Neutral, Disgust, Surprise, Boredom |

**Table A.6**
Devices, sensors, and specific models used in the reviewed studies.

| Manuscript | Sensor | Device | Model |
| --- | --- | --- | --- |
| Piana et al. (2016) [48] | Vision sensor, Pose tracking sensor | Kinect | Kinect / Kinect 2, Qualisys |
| Irani et al. (2018) [8] | Not sufficiently described | Tablet | Not sufficiently described |
| Leo et al. (2015) [49] | Vision sensor | Robot | R25 robot from Robokind |
| Postawka et al. (2019) [17] | Vision sensor | Kinect | Kinect |
| Jiang et al. (2019) [18] | Vision sensor | Head and eye tracker | Tobii Pro TX300, Tobii X2-60 |
| Gao et al. (2015) [19] | EEG sensor | Brain–Computer Interface | Emotiv EPOC neuroheadset |
| Heni et al. (2016) [35] | Vision sensor | Mobile phone | Camera from mobile phone |
| Jeon et al. (2015) [50] | Vision sensor | Mobile phone, Robot | ROMO robot |
| Fan et al. (2017) [46] | ECG sensor | Brain–Computer Interface | Emotiv EPOC neuroheadset |
| Joseph et al. (2018) [20] | Vision sensor | Raspberry Pi | Camera undefined |
| Spicker et al. (2016) [79] | Handy actuator | Keypad | Keypad undefined |
| Santhoshkumar et al. (2019) [21] | Vision sensor | Computer | Camera undefined |
| Tang et al. (2017) [51] | Vision sensor | 3D Camera | Intel RealSenseTM SR 300 |
| Ley et al. (2019) [62] | Vision sensor, Audio sensor | Computer | Camera undefined, Microphone undefined |
| Chung et al. (2019) [52] | Vision sensor | Augmented reality sensor | Microsoft Hololens |
| Smitha et al. (2015) [53] | Not sufficiently described | IoT board | Virtex 7 XC7VX330T FFG1761-3 |
| Daniels et al. (2018) [54] | Vision sensor | Augmented reality sensor | Google Glass |
| Smitha et al. (2013) [55] | Not sufficiently described | IoT board | Virtex 7 XC7VX330T FFG1761-3 FPGA |
| Tang et al. (2016) [56] | Vision sensor, Wearable sensor, Capacitive sensor | IP camera, IoT board | Kinect, Microsoft Band 2, IP camera |
| Liliana et al. (2020) [23] | Vision sensor | Camera | Panasonic AG-7500 cameras |
| Ghorbandaei et al. (2018) [72] | Vision sensor | Robot, Kinect | R50-Alice by Hanson RoboKind Company, Kinect |
| Fernandes et al. (2011) [36] | Vision sensor | Computer | Camera undefined |
| Anishchenko et al. (2017) [37] | Vision sensor | Tablet, Computer | iPad, Web camera undefined |
| Su et al. (2018) [78] | Vision actuator | Tablet, Eye tracker, Computer | Samsung ST 800, Binocular flat screen |
| Arellano et al. (2015) [75] | Vision sensor | Eye tracker, Computer | RED250, Web camera undefined |
| AndleebSiddiqui et al. (2020) [25] | Vision sensor, Audio actuator, Audio sensor | Microphone | Camera undefined, Speaker undefined, Microphone undefined |
| Globerson et al. (2012) [77] | Audio actuator, Audio sensor | Headphone, Microphone | Headphone, Recorder device |
| Sunitha et al. (2014) [73] | Vision sensor, Audio sensor | Computer | Web camera undefined, Microphone undefined |
| Bagirathan et al. (2020) [47] | ECG sensor, Electrodes | Computer, TV | Shimmer device for ECG data |
| Marchi et al. (2015) [74] | Audio sensor, Handy actuator | Computer | Microphone undefined |
| Syeda et al. (2017) [57] | Vision sensor | Eye tracker | Tobii EyeX |
| Guha et al. (2018) [76] | Vision sensor | Infrared motion captured camera | Not sufficiently described |
| Costescu et al. (2020) [58] | Not sufficiently described | Tablet | Not sufficiently described |
| Tracy et al. (2011) [80] | Not sufficiently described | Computer, TV | Not sufficiently described, Not sufficiently described |
| Chung et al. (2020) [52] | Vision sensor | Computer | Digital/web camera |
| Zhang et al. (2016) [60] | Vision sensor | Computer | Digital camera Canon |
| Dantas et al. (2022) [38] | Vision sensor | Webcam | Not sufficiently described |
| Saranya et al. (2022) [28] | Vision sensor | Not sufficiently described | Not sufficiently described |
| Sukumaran et al. (2021) [63] | Audio sensor | Microphone | Not sufficiently described |
| Wang et al. (2021) [30] | Vision sensor | Mobile phone | Huawei G9 VNS-AL00 smartphone |
| Banire et al. (2021) [29] | Vision sensor | Computer, camera | Logitech web camera |
| Piana et al. (2021) [39] | Vision sensor | RGB-D sensor | Not sufficiently described |
| Ruan et al. (2022) [64] | Vision sensor, light sensor | Camera, light | Panasonic HPX 370, 3200 Soft Light |
| Milling et al. (2022) [65] | Vision sensor, audio sensor | Camera, Microphone | Not sufficiently described |
| Chitre et al. (2022) [66] | Audio sensor | Microphone | Not sufficiently described |
| Wang et al. (2022) [67] | Vision sensor | Computer, camera | Not sufficiently described |
| Wan et al. (2022) [40] | Vision sensor | Computer, mobile phone | Not sufficiently described |
| Silva et al. (2021) [41] | Vision sensor, robot | Camera, computer, robot | Intel RealSense sensor model F200, ZECA robot (Zeno R50) |
| Praveena et al. (2021) [68] | Vision sensor | Not sufficiently described | Not sufficiently described |
| Rojas et al. (2021) [42] | Vision sensor | Not sufficiently described | Not sufficiently described |
| Karanchery et al. (2021) [43] | Vision sensor | Camera | Not sufficiently described |
| Valles et al. (2021) [69] | Not sufficiently described | Not sufficiently described | Not sufficiently described |
| DIzicheh et al. (2021) [44] | Not sufficiently described | Computer | Not sufficiently described |
| Pulido-Castro et al. (2021) [31] | Not sufficiently described | Computer, Nvidia GeForce | Windows 10 64 bits OS, Intel Core i7-6700HQ processor, 16GB of RAM; Nvidia GeForce GTX and 1060 6GB GPU |
| Arabian et al. (2021) [32] | Not sufficiently described | Not sufficiently described | Not sufficiently described |
| Li et al. (2021) [33] | Not sufficiently described | Not sufficiently described | Not sufficiently described |
| Ghanouni et al. (2021) [45] | Vision sensor | Kinect | Not sufficiently described |
| Zhang et al. (2023) [70] | Vision | Camera | Not sufficiently described |
| Talaat (2023) [34] | Not sufficiently described | Not sufficiently described | Not sufficiently described |
| Murugaiyan et al. (2023) [71] | Audio | Microphone | Not sufficiently described |

Table A.7
Machine learning techniques, performance and metrics, and validation methods used in each of the reviewed studies.

| Manuscript | ML techniques | Validation type | Performance metric | Average performance |
|---|---|---|---|---|
| Piana et al. (2016) [48] | SVM, Sparse Dictionary Learning | Random split - leave one subject out | Accuracy | 86.40 |
| Irani et al. (2018) [8] | SVM, DT, LR, KNN | Not sufficiently described | Accuracy | 91.03 |
| Leo et al. (2015) [49] | Histograms of oriented gradients (HOG), SVM | 10 cross-validation | Accuracy | 98.90 |
| Postawka et al. (2019) [17] | HMM | Not sufficiently described | Accuracy | 78.96 |
| Jiang et al. (2019) [18] | RF | one-leave cross-validation | accuracy, sensitivity and specificity | 86.20 |
| Gao et al. (2015) [19] | RBM, KNN, SVM, ANN | 10 cross-validation | Accuracy | 68.40 |
| Heni et al. (2016) [35] | Face Emotion Recognizer | Not sufficiently described | Accuracy | 87.00 |
| Jeon et al. (2015) [50] | SVM | Not sufficiently described | Accuracy | 58.00 |
| Fan et al. (2017) [46] | Bayesian networks, SVM, ANN, kNN, RF, DT | 5 and 8 cross-validation | Accuracy | 82.50 |
| Joseph et al. (2018) [20] | Deep learning networks | Not sufficiently described | Accuracy | 67.50 |
| Spicker et al. (2016) [79] | Does not apply | Not sufficiently described | Not sufficiently described | Not sufficiently described |
| Enticott et al. (2014) [61] | LR | Not sufficiently described | Not sufficiently described | Not sufficiently described |
| Santhoshkumar et al. (2019) [21] | RF, SVM | 10 cross-validation | Accuracy | 96.30 |
| Sivasangari et al. (2019) [22] | ANN | Not sufficiently described | accuracy, sensitivity and specificity | 86.86 |
| Tang et al. (2017) [51] | RealSense Facial Recognition | qualitative analysis | Sensitivity | 65.00 |
| Ley et al. (2019) [62] | FaceReader | Not sufficiently described | Accuracy | 45.50 |
| Chung et al. (2019) [52] | CNN | random validation | Accuracy | 69.50 |
| Smitha et al. (2015) [53] | PCA | 10 cross-validation | Accuracy | 82.30 |
| Daniels et al. (2018) [54] | LR | 10 cross-validation | Accuracy | 72.70 |
| Smitha et al. (2013) [55] | PCA | Not sufficiently described | Accuracy | 72.60 |
| Tang et al. (2016) [56] | Does not apply | Surveys | Not sufficiently described | Not sufficiently described |
| Liliana et al. (2020) [23] | FUZZY, SVM | Not sufficiently described | Accuracy | 98.26 |
| Ghorbandaei et al. (2018) [72] | FuzzyC-Means (FCM) | Not sufficiently described | Sensitivity | 93.20 |
| Elamir et al. (2018) [24] | RQA, SVM, KNN, RF | Not sufficiently described | Accuracy | 93.15 |
| Fernandes et al. (2011) [36] | Not provided | qualitative analysis | results expressed qualitatively | Not sufficiently described |
| Anishchenko et al. (2017) [37] | Not provided | Not sufficiently described | | Not sufficiently described |
| Su et al. (2018) [78] | Not provided | Not sufficiently described | Accuracy | 81.00 |
| Arellano et al. (2015) [75] | Not provided | Not sufficiently described | Accuracy | 61.80 |
| AndleebSiddiqui et al. (2020) [25] | Deep learning networks | Not sufficiently described | Accuracy | 46.50 |
| Globerson et al. (2012) [77] | LR, Bivariate correlation | Not sufficiently described | Accuracy | 35.70 |
| Sunitha et al. (2014) [73] | SVM | Not sufficiently described | Accuracy | 80.00 |
| Bagirathan et al. (2020) [47] | SVM, KNN, Ensemble classifiers | Not sufficiently described | Accuracy | 85.50 |
| Piparsaniyan et al. (2014) [26] | Bayesian networks | Split test-train | Accuracy | 96.73 |
| Marchi et al. (2012) [27] | SVM | Not sufficiently described | UAR | 93.05 |
| Marchi et al. (2015) [74] | SVM | one leave cross-validation | UAR | 82.40 |
| Syeda et al. (2017) [57] | Not provided | Qualitative analysis | Not sufficiently described | Not sufficiently described |
| Guha et al. (2018) [76] | MSE | Not sufficiently described | Accuracy | 90.00 |
| Costescu et al. (2020) [58] | Not provided | Not sufficiently described | Accuracy | 87.00 |
| Tracy et al. (2011) [80] | Not provided | Not sufficiently described | Accuracy | 89.00 |
| Chung et al. (2019) [52] | Not provided | Not sufficiently described | Accuracy | 93.34 |
| Zhang et al. (2016) [60] | Fuzzy, SVM | 10 cross-validation | Accuracy | 85.00 |
| Dantas et al. (2022) [38] | Decision trees | Hold out | Accuracy | 88.84 |
| Saranya et al. (2022) [28] | Deep learning networks | Leave one out cross-validation | Accuracy | 92.50 |
| Sukumaran et al. (2021) [63] | MLP Classifier | Split test-train | Accuracy | 81.52 |
| Wang et al. (2021) [30] | CNN | Not sufficiently described | Accuracy, precision, recall, F1-score | 95.89, 97.58, 100, 98.79 |
| Banire et al. (2021) [29] | SVM, CNN | 10-cross validation | Accuracy, AUC | 88.9, 53.1 |
| Piana et al. (2021) [39] | Linear SVM | Surveys | Accuracy | 64.48 |
| Ruan et al. (2022) [64] | TL, NN | Not sufficiently described | Accuracy | 30.0 |
| Milling et al. (2022) [65] | LSTM | Not sufficiently described | ROC-AUC, CCC, RMSE | 75.6, 26.3, 10.7 |
| Chitre et al. (2022) [66] | CNN | Not sufficiently described | Accuracy | 89.93 |
| Wang et al. (2022) [67] | Does not apply | Surveys | Accuracy | 98.00 |
| Wan et al. (2022) [40] | CNN | Not sufficiently described | Accuracy | 74.78 |
| Silva et al. (2021) [41] | SVM, RBF | Not sufficiently described | Accuracy, Sensitivity, Specificity, AUC, MCC | 91.1, 92.85, 97.9, 98.1, 88.25 |
| Praveena et al. (2021) [68] | HCC, NN | Split test-train | Accuracy | 40.0 |
| Rojas et al. (2021) [42] | HCC, CNN | Split test-train | Accuracy | 84.0 |
| Karanchery et al. (2021) [43] | SNN, NN | Split validation | Accuracy | 99.8 |
| Valles et al. (2021) [69] | SVM, MLP, RNN | Split test-train | Accuracy | 65.5 |
| DIzicheh et al. (2021) [44] | Does not apply | Does not apply | Does not apply | Does not apply |
| Pulido-Castro et al. (2021) [31] | ANNs, SVMs, KNN, RFs | 10-fold cross-validation | Accuracy | 71.0, 73.0 |
| Arabian et al. (2021) [32] | Does not apply | Cross validation | Accuracy | 99.9 |
| Li et al. (2021) [33] | MTCNN | 5-fold cross-validation | Accuracy | 72.40 |
| Ghanouni et al. (2021) [45] | Does not apply | Does not apply | Does not apply | Does not apply |
| Zhang et al. (2023) [70] | Own method | Not sufficiently described | Accuracy | 91.72 |
| Talaat (2023) [34] | CNN | Not sufficiently described | Does not apply | Does not apply |
| Murugaiyan et al. (2023) [71] | CNN, LSTM | Not sufficiently described | Does not apply | Does not apply |

Table A.8

Information privacy and security policies considered in those reviewed studies acknowledging these aspects.

| Manuscript | Information privacy and security |
| --- | --- |
| Jian et al. (2019) [18] | All subjects were recruited with the approval of the University of Minnesota (UMN) Institutional Review Board. |
| Spicker et al. (2016) [79] | The study protocol was performed in accordance with the ethical standards laid down in the Declaration of Helsinki and its later amendments. |
| Daniels et al. (2018) [54] | Participants' assent and parents' informed consent were received before inclusion in the study. |
| Liliana et al. (2020) [23] | Participants and parents provided written informed consent under an approved Stanford University IRB protocol, which followed the guidelines of the Declaration of Helsinki prior to their inclusion in the study (the parent-completed Social Responsiveness Scale-2, SRS-253, were collected. |
| Ghorbandaei et al. (2018) [72] | Participants and their parents signed a consent form for moral obligations. All procedures performed in studies were in accordance with the ethical standards of the institutional and/or national research committee and with the Declaration of Helsinki declaration and its later amendments or comparable ethical standards. |
| Elamir et al. (2018) [24] | Participants and their parents signed a consent form for moral obligations. |
| Fernandes et al. (2011) [36] | The consents of the parents of the participants were collected. |
| Bagirathan et al. (2020) [47] | Ethical approval was obtained from the Ethics Committee of the National Institute for Empowerment of Persons with Multiple Disabilities (NIEPMD). The consents of the parents of the participants were collected. |
| Costescu et al. (2020) [46] | Ethical approval was obtained from the Ethics Committee of Babes-Bolyai University. All the participants' parents signed an informed consent to participate in the study. |
| Enicot et al. (2014) [61] | Ethical approval was obtained from the human research ethics committees of Monash University and The Alfred hospital (Bayside Health). |
| Banire et al. (2021) [29] | Approval was obtained from the institutional review board. |
| Piana et al. (2021) [39] | The consents of the parents of the participants were collected. |
| Wang et al. (2022) [67] | The choice of a multiple baseline design over a reversal design was based on ethical concerns given the ability of emotion recognition ability by participants during the intervention phase. |
| Silva et al. (2021) [41] | Approval of the Ethics Committee of the University and Informed Consents from children's parents or those responsible were obtained prior to the experiments. |